\documentclass[preprint,3p]{elsarticle}

\usepackage{amssymb}
\usepackage{amsmath}
\usepackage{hyperref}
\usepackage{array}
\usepackage{multirow}
\usepackage{multicol}
\usepackage{subcaption}
\usepackage{nomencl}
\makenomenclature
\usepackage{framed} 
\usepackage{xcolor}
\usepackage{comment}
\usepackage{booktabs}
\usepackage{orcidlink}
\usepackage{float}

\renewcommand*\nompreamble{\begin{multicols}{2}}
\renewcommand*\nompostamble{\end{multicols}}

\journal{}

\makeatletter
\def\ps@pprintTitle{%
    \let\@oddhead\@empty
    \let\@evenhead\@empty
    \let\@oddfoot\@empty
    \let\@evenfoot\@empty
}
\makeatother

\begin{document}

\begin{frontmatter}




\title{A Trustworthy By Design Classification Model for Building Energy Retrofit Decision Support}


\author{Panagiota Rempi\corref{cor1}\fnref{label1}\orcidlink{0009-0003-6628-8131}}
\ead{prempi@epu.ntua.gr}
\cortext[cor1]{Corresponding author}
\author[label1]{Sotiris Pelekis\orcidlink{0000-0002-9259-9115}}
\ead{spelekis@epu.ntua.gr}
\author[label1]{Alexandros Menelaos Tzortzis\orcidlink{0009-0002-5020-4485}}
\ead{atzortzis@epu.ntua.gr}
\author[label1]{Evangelos Spiliotis\orcidlink{0000-0002-1854-1206}}
\ead{vspiliotis@epu.ntua.gr}
\author[label1]{Evangelos Karakolis\orcidlink{0000-0003-2833-3088}}
\ead{vkarakolis@epu.ntua.gr}
\author[label1]{Christos Ntanos\orcidlink{0000-0002-5162-6500}}
\ead{cntanos@epu.ntua.gr}
\author[label1]{Dimitris Askounis\orcidlink{0000-0002-2618-5715}}
\ead{askous@epu.ntua.gr}

\affiliation[label1]{organization={Decision Support Systems Laboratory, School of Electrical and Computer Engineering, National Technical University of Athens},
             addressline={Iroon Politechneiou 9}, 
             city={Zografou},
             postcode={15772}, 
             state={Attica},
             country={Greece}}

\begin{abstract}

Improving energy efficiency in residential buildings is critical to combating climate change and reducing greenhouse gas emissions. Retrofitting existing buildings, which contribute a significant share of energy use, is therefore a key priority, particularly in regions with outdated building stock. Artificial Intelligence (AI) and Machine Learning (ML) can automate retrofit decision-making and find retrofit strategies. However, their implementation faces challenges of data availability, model transparency, and compliance with national and EU AI regulations including the AI act, ethics guidelines and the ALTAI. 

This paper presents a trustworthy-by-design ML-based decision support framework that recommends energy efficiency strategies for residential buildings using minimal user-accessible inputs. The framework merges Conditional Tabular Generative Adversarial Networks (CTGAN) to augment limited and imbalanced data, while a neural network-based multi-label classifier identifies potential combinations of retrofit measures. To support explanation and trustworthiness, an Explainable AI (XAI) layer using SHapley Additive exPlanations (SHAP) is incorporated to clarify the rationale behind recommendations and guide feature engineering.

Two case studies validate performance and generalization: the first leveraging a well-established, large EPC dataset for England and Wales; the second using a small, imbalanced post-retrofit dataset from Latvia (RETROFIT-LAT). Results show that the framework can handle diverse data conditions and improve performance up to 53\% compared to the baseline, confirming its effectiveness. Overall, the proposed framework provides a feasible, interpretable, and trustworthy AI system for building retrofit decision support through assured performance, usability, and transparency to aid stakeholders in prioritizing effective energy investments and support regulation-compliant, data-driven innovation in sustainable energy transition.

\end{abstract}




\end{frontmatter}



\section{Introduction} \label{intro}

\begin{table*}[h]   
\begin{framed}
\nomenclature{AI}{Artificial intelligence}
\nomenclature{XAI}{Explainable artificial intelligence}
\nomenclature{ML}{Machine learning}
\nomenclature{DL}{Deep learning}
\nomenclature{EPC}{Energy Performance Certificate}
\nomenclature{GHG}{Greenhouse gas}
\nomenclature{MLP}{Multilayer perceptron}
\nomenclature{SHAP}{SHapley Additive exPlanations}
\nomenclature{EU}{European Union}
\nomenclature{EPBD}{European Union Energy Performance of Buildings Directive}
\nomenclature{CO$_2$}{Carbon dioxide}
\nomenclature{BCE}{Binary cross-entropy}
\nomenclature{TPE}{Tree-structured parzen estimator}
\nomenclature{CTGAN}{Conditional Tabular Generative Adversarial Network}
\nomenclature{GAN}{Generative Adversarial Network}
\nomenclature{SDV}{Synthetic Data Vault}
\nomenclature{TN}{True negatives}
\nomenclature{TP}{True positives}
\nomenclature{FN}{False negatives}
\nomenclature{FP}{False positives}
\nomenclature{ReLU}{Rectified linear unit}

\printnomenclature
\end{framed}
\end{table*}

\subsection{Background}

With the growing energy demand and the intensifying effects of climate change, energy efficiency has emerged as a critical objective worldwide. The Paris Agreement underscores the international commitment to limit global warming well below 2°C above pre-industrial levels with efforts to restrict it to 1.5°C \cite{ParisAgreement}. Towards this target, the building sector has become one of the priorities, since it is among the largest energy consumers and greenhouse gas (GHG) emitters. Specifically, in the European Union (EU) buildings account for around 40\% of the total energy consumption and over 30\% of GHG emissions. Notably, 85\% of buildings were built before 2000, and 75\% of these are considered energy-inefficient due to outdated construction standards \cite{EnergyEC}. In this context, EU has established the Energy Performance of Buildings Directive (EPBD) with its latest revision in 2024 \cite{EPBD24}. This legislative framework acts as a part of the strategy to reduce building sector's GHG emissions by 2030 and achieve a decarbonized building stock by 2050 \cite{Economidou2020}.

Residential buildings constitute the majority of structures across EU with the largest energy demand and environmental impact among different building types \cite{BSO}. Approximately 80\% of residential energy consumption is used for heating, cooling, and hot water, highlighting significant potential for improvements \cite{EnergyEC}. Therefore, the construction of new energy efficient buildings is not sufficient and retrofitting of existing buildings is also urgent to mitigate their environmental footprint \cite{Economidou2020}. 

The implementation of appropriate renovation measures in the dwellings is a promising solution to improve their energy efficiency \cite{Ma2012}. This can be achieved through various technologies and practices, such as enhancing insulation, upgrading heating systems, integrating renewable energy sources \cite{KOUTALIDIS2025380, Pelekis2023} etc. In addition to energy-savings, other benefits also include reduced maintenance costs, increased property values and enhanced occupant comfort, health and well-being \cite{Ma2012}. Likewise, it leads to lower energy bills alleviating energy poverty which affects vulnerable consumers \cite{EnergyEC}.

It should be noted that the retrofitting process is considered a complex task that demands careful planning and evaluation of multiple factors \cite{Ma2012}. These challenges are highlighted by the low renovation rate, which ranges from 0.4\% to 1.2\% across EU Member States \cite{Fetting2020}. Therefore, it is evident that the development of methods to assist building owners, investors, and governments in selecting optimal retrofit solutions and estimating their benefits before implementation is crucial. Traditionally, such decisions are based on energy audits and analysis by domain specialists to estimate the potential of each retrofit measure. However, these methods are often impractical to apply on a large scale due to their extensive data and resource requirements. Energy audits demand comprehensive building-specific information, onsite inspections and expert evaluation, which entail considerable time and financial costs. Accordingly, research focuses on developing a variety of techniques to facilitate the decision-making process, including engineering and data-driven models. Artificial intelligence (AI), particularly machine learning (ML) and deep learning (DL), which have been increasingly applied to the built environment sector \cite{Tien2022}, is a very promising approach for building retrofit, as it can extrapolate knowledge from previous renovations and generalize data relationships across buildings at minimal cost. These algorithms are highly effective at analyzing the complexities of energy data, revealing hidden trends and interdependencies \cite{Alotaibi2024} and thus playing a major role in enhancing retrofit decision-making.


Despite their capabilities, the successful application of such models is heavily dependent on the availability of high-quality data \cite{Yilmaz2022}. ML and DL algorithms require substantial and representative datasets to learn meaningful patterns and make accurate predictions \cite{aditya_sai_srinivas_2023}. Insufficient data can lead to poor model fitting, limited generalization, and unreliable results, particularly in specialized domains like energy retrofitting where relationships between building characteristics, energy consumption and retrofit options must be captured. In such scenarios, datasets are often scarce, inconsistent, or incomplete, limiting the effectiveness of the models \cite{Fan2022}. Therefore, addressing these data challenges is essential and could be accomplished through techniques, such as synthetic data generation, which create diverse and realistic samples to augment the initial datasets \cite{lu2024machinelearningsyntheticdata}. Additionally, leveraging feature engineering methods has the potential to further optimize the use of limited data and enhance the performance of the developed models \cite{Fan2019}.


Another issue associated with the application of advanced AI models is the need to ensure trustworthiness and regulatory compliance alongside predictive performance \cite{Arrieta2020}, with the energy sector gradually gaining significant interest \cite{10794222}. While DL architectures such as multi-layer perceptrons (MLPs) are highly effective, their “black-box” nature limits interpretability and transparency, raising concerns about accountability and user trust \cite{Pelekis2025}. In response, explainable artificial intelligence (XAI) has emerged as a means to render the reasoning of these systems understandable and verifiable, supporting human oversight and informed decision-making \cite{Machlev2022, Adadi2018}. Within the European context, this direction is reinforced by the forthcoming \textit{AI Act} \cite{EC_AIAct2025}, expected to enter into force in 2026, which defines requirements for transparency, human supervision, robustness, and data quality that are criteria particularly relevant to AI applications in the energy and buildings sector. Embedding these trustworthy-by-design principles from the early development stage ensures that AI systems are not only accurate but also compliant with ethical and legal standards such as the \textit{Ethics Guidelines for Trustworthy AI} and the \textit{ALTAI} framework \cite{EC_ALTAI2020}, in turn increasing trust, reducing administrative burden and promoting adoption by stakeholders such as policy makers, funding agencies, and building owners.

\subsection{Objective of the study}
The purpose of this study is to introduce a decision support framework based on multi-label classification to recommend suitable energy efficiency retrofit measures for residential buildings. The framework is designed for homeowners and investors, while prioritizing simplicity in user data submission, by only requiring a minimal set of inputs \citep{TZORTZIS2025102172} such as basic building characteristics, location data, and information derived from energy performance certificates (EPCs). Overall, the model suggests the proper combination of retrofit actions to achieve a target energy class specified by the user.

The design of the decision support framework is based on trustworthiness-by-design principles and is centered around the requirements of the recently established AI act \cite{EC_AIAct2025} aiming to address compliance to upcoming regulations. In this direction, we employ domain knowledge and XAI to derive proper investment input features. Besides feature selection, the XAI analysis provides model explanations to increase the framework's transparency and interpretability. Additionally, aiming to address the common limitations of energy retrofitting datasets--such as data scarcity and class-imbalance--we propose an additional synthetic data generation layer based on GANs.

Our framework is first validated on the case study of the UK, using the Energy Performance of Buildings Data: England and Wales'' dataset \cite{englishEPCs}, which provides a rich yet class-imbalanced source of information on building energy performance. To further evaluate our approach under data scarcity, we also apply it to the RETROFIT-LAT'' dataset \cite{LEIF_DATA_PAPER}, which contains information on Latvian private houses that improved their energy efficiency through renovation projects.

\subsection{Structure of the paper}
The rest of the paper is structured as follows. Section \ref{relatedwork} provides an overview of the existing literature relevant to our research and highlights the key contributions of this study. Section \ref{method} describes our proposed modelling methodology and its trustworthy AI-centered architecture along with our experimental setup. Sections \ref{case1} and \ref{case2} present the two case studies conducted, describing the datasets used and the corresponding results. Section \ref{conclusions} concludes the study, including a comparative discussion of the model's results and also stating potential future extensions. 

\section{Literature review and contributions} \label{relatedwork}
\subsection{Literature review}
\subsubsection{Building retrofitting}
\paragraph{Energy simulations and statistical models}
Improving the energy efficiency of buildings by applying appropriate energy retrofit measures has been studied from various perspectives using diverse methods. Research has primarily focused on energy simulations employing tools such as EnergyPlus, which are often combined with other models, to evaluate potential retrofit scenarios in terms of energy, environmental, and financial aspects \cite{Pasichnyi2019}. The energy performance of a building is calculated and possible measures are assessed by comparing the results of multiple tests with varying input parameters. Similarly, computational statistical models have been developed to estimate energy savings achieved by different combinations of measures. They rely on predefined physical equations using buildings characteristics to draw conclusions about the effectiveness of each option, thereby facilitating the decision-making process \cite{Uidhir2020, Gouveia2019, Mutani2023}. However, despite the reliability and transparency of these methods, their computational load can often be prohibitive. Additionally, the large number of required input parameters poses another limitation, as detailed information, such as construction details and thermal coefficients, is often unavailable, particularly for older buildings \cite{Deb2021}.

\paragraph{Clustering algorithms and benchmarking}
Clustering algorithms are also frequently encountered in the literature, particularly in cases where post-retrofit performance data is unavailable. Buildings are grouped based on their similar characteristics and serve as benchmarks in order to identify potential energy efficiency measures \cite{Ali2020, Deb2021, Penaka2023}. For example, Cecconi et al.\cite{ReCecconi2022} cluster buildings per energy class and apply Monte Carlo simulations to estimate energy savings for various retrofit options within each class. For the final selection they use a decision support system which considers cost and average energy savings.  

\paragraph{Multi-criteria analysis and optimization approaches}
Multi-criteria analysis and optimization have been widely applied to assess retrofit scenarios regarding multiple environmental, financial and practical factors. For instance, Asadi et al. \cite{Asadi2012} propose a multi-objective optimization model to identify cost-effective interventions that minimize energy use while meeting occupant needs. A cost-benefit analysis is conducted by Belaïd et al. \cite{Belaïd2021} to examine both environmental and economic indicators. Multi-objective optimization is also often used with genetic algorithms, such as NSGA-II \cite{Penna2015}, but these approaches come with limitations including complexity and the need for a high level of expertise \cite{CostaCarrapiço}. 

\paragraph{Pure machine learning approaches}
Recently, ML has drawn considerable attention in the field of building retrofitting due to its ability to reduce both the computational cost and the number of parameters required from users. Such data-driven models can identify complex relationships between building features and predict energy performance using only a limited set of basic characteristics. However, their effective implementation requires large and diverse datasets. Based on our research and as noted in \cite{Deb2021} no studies provide direct energy retrofit recommendations by exclusively applying ML models on measured datasets.

\paragraph{Energy consumption prediction}
Several publications approach the problem by predicting the final energy consumption after implementing an energy efficiency measure. In this context, regression models such as neural networks take building parameters and potential measures as input and predict the associated energy savings, which are then compared to one another \cite{Walter2016}. The output may also involve additional metrics, including cost or $\text{CO}_2$ emissions reduction \cite{Feng2022}. Alternatively, post-retrofit energy class of a building is predicted instead, as in \cite{Sun2022}. Similarly, Seraj et al. \cite{Seraj2024} develop multiple ML models to predict annual energy consumption, and a user interface which allows stakeholders to simulate and evaluate the impact of different renovation scenarios based on the specific characteristics of their dwellings. Finally, Michalakopoulos et al. \cite{Michalakopoulos2023} introduce a physics-informed neural network --integrated withing a decision support system \cite{Michalakopoulos2025}-- that incorporates physics equations into the loss function to predict energy consumption based on heat losses, while identifying specific building components requiring retrofit actions.

\paragraph{Hybrid approaches}
The integration of ML with other methods is often proposed. Cecconi et al. \cite{Cecconi2023} combine ML models with Monte Carlo simulations to predict energy savings from retrofit scenarios and estimate their costs. ML is combined with multi-criteria optimization by Araújo et al. \cite{Araújo2024}. Regression models, which are trained to predict energy needs and energy label are followed by optimization to explore the best retrofit strategies depending on users' criteria. Deb et al. \cite{Deb2021ML} predict energy demand with ML models and then with cost-optimal analysis they identify potential retrofit options. A hybrid model is proposed by Long \cite{Long2023}, integrating simulation, ML for consumption prediction, and optimization models to evaluate energy strategies even during the early design stages of a building.

\paragraph{System-level control and economic assessment}
Ultimately, a significant share of research is shifted towards controlling the energy use of individual systems such as heating and ventilation. Haidar et al. \cite{Haidar2023} employ reinforcement learning to predict occupant behavior within building spaces, using the results to optimize the operation of energy systems, while Homod et al. \cite{Homod2023} develop a deep clustering multi-agent cooperative reinforcement learning method to manage the operation of cooling systems in buildings, achieving energy savings. ML has also been applied to address the economical aspects of energy retrofits, assessing energy efficiency investments. Sarmas et al. \cite{Sarmas2022} label these investments based on renovation costs and energy savings combining ML models through a meta-learning model determining the potential funding for each solution. 

\subsubsection{Trustworthy and explainable artificial intelligence}
Recently, several studies have integrated ML models with XAI techniques to enhance transparency and derive actionable insights for energy savings. These models predict various aspects of building energy efficiency, such as energy consumption \cite{Cui2024, Liu2022}, energy efficiency ratings \cite{Sun2022}, heating and cooling loads \cite{Alotaibi2024}, GHG emissions \cite{Zhang2023}. 
Explainability methods, particularly Shapley additive explanations (SHAP) \cite{NIPS2017_8a20a862}, are then applied to identify the most influential features affecting these predictions. By quantifying the impact of these features, this analysis enables the prioritization of retrofitting directions, focusing on the elements that most significantly contribute to building performance. Similarly, Deb et al. \cite{Deb2021ML} propose a ML-based framework for cost-optimal retrofit analysis, leveraging feature significance through SHAP to guide retrofit strategies. For instance, better materials are assigned to critical building areas determined by their influence on heating demand, thereby eliminating the need for exhaustive search methods required in conventional approaches. In most cases, only a discussion of potential retrofit strategies is conducted, without explicit recommendations or detailed analysis.

XAI has also been employed to assess the potential economic benefits of energy efficiency improvements in the building sector. A cluster-based XAI methodology is proposed by Konhäuser and Werner \cite{Konhäuser}, combining SHAP, Permutation Feature Importance, and Partial Dependency Plots to reveal the financial impacts of energy-efficient building features on property valuation. Wenninger et al. \cite{Wenninger2022} incorporate SHAP with a XGBoost model to predict whether buildings had undergone retrofits, identifying key factors influencing retrofit decisions. Policy recommendations for renovation programs were derived by the investigation of building characteristics, house prices and socio-demographic data.

Although SHAP is the predominant XAI method in this field, other tools have been explored as well. Nyawa et al. \cite{Nyawa2023} base their research on sensitivity analysis to assess variable importance in predicting retrofit decisions (i.e. whether retrofitting was implemented, not specific measures). The feature importance measure was determined by accuracy reduction upon feature removal, with high accuracy impact indicating great significance for renovation decisions. Collectively, the review of XAI applications in retrofitting highlights their value not only in improving model explainability but also in providing actionable insights to guide the evaluation and prioritization of retrofit measures.

\subsection{Paper contributions} 
The contributions of this study are summarized as follows:
\begin{itemize}

\item We introduce a novel decision support framework based on neural networks and multi-label classification. The framework facilitates deep renovation by directly recommending combinations of retrofitting strategies in a single step, thanks to its multi-label nature.  This contrasts with previous studies \cite{Walter2016, Sun2022, Seraj2024, Deb2021ML} that rely on complex and computationally intensive simulations, optimization models, or single-measure predictions. These typically produce isolated recommendations requiring additional manual or computational effort to compare alternatives and assemble complete renovation packages that account for the interdependencies among measures \cite{Ma2012}. 

\item To enhance usability for non-expert users such as household owners, the framework only requires basic building characteristics as input and outputs recommendations at the level of umbrella categories (e.g., building fabric interventions, heating \& lighting controls, DHW upgrades, and heating system installation) rather than the full feature space of highly specific measures. This abstraction reduces cognitive load, aligns with EPC and EU retrofit communication practices, and ensures interpretability and actionability for end-users.

\item We propose a trustworthy-by-design framework by integrating XAI, and GAN-based synthetic data generation. XAI optimizes the feature space and explains retrofit recommendations, while GANs mitigate class imbalance and data scarcity \cite{Baset2024}. Together, they amplify the value of real data, strengthen data-driven decision-making, and align with regulatory requirements on explainability, oversight, and data quality. This design approach can be leveraged by funding agencies, investors, and policy makers to develop AI systems within the EU energy efficiency investment landscape that are compliant with the upcoming AI Act and related ethical guidelines, such as the ALTAI.

\item Our design approach uses XAI, and specifically SHAP, to directly explain the discrete retrofit recommendations of the classifier, unlike most previous studies that apply explainability merely to assess lower-level energy efficiency factors, such as energy consumption \cite{Cui2024, Sun2022, Alotaibi2024, Zhang2023} or simple yes or no retrofit decisions \cite{Wenninger2022, Nyawa2023} (i.e., whether any retrofit measures were applied, without specifying which one). 

\item We validate our decision support framework on two case studies: i) the well-established UK EPC dataset \cite{englishEPCs}, which establishes the generalizability of our methodology; ii) the real-world post-retrofit audit dataset from Latvia (RETROFIT-LAT dataset \cite{LEIF_DATA_PAPER}), whose limited size and class imbalance allow us to examine how the proposed trustworthy-by-design framework mitigates these challenges, while also addressing the energy efficiency gap \cite{Backlund2012, Jaffe1994}, in contrast to most previous studies that solely rely on ex-ante EPC estimates \cite{Backlund2012, Jaffe1994, Walter2016, Sun2022, Ali2020}. 

\end{itemize}

\section{Methodology and experimental setup} \label{method}

\begin{figure}[htbp]
\centering
\includegraphics[scale=0.33]{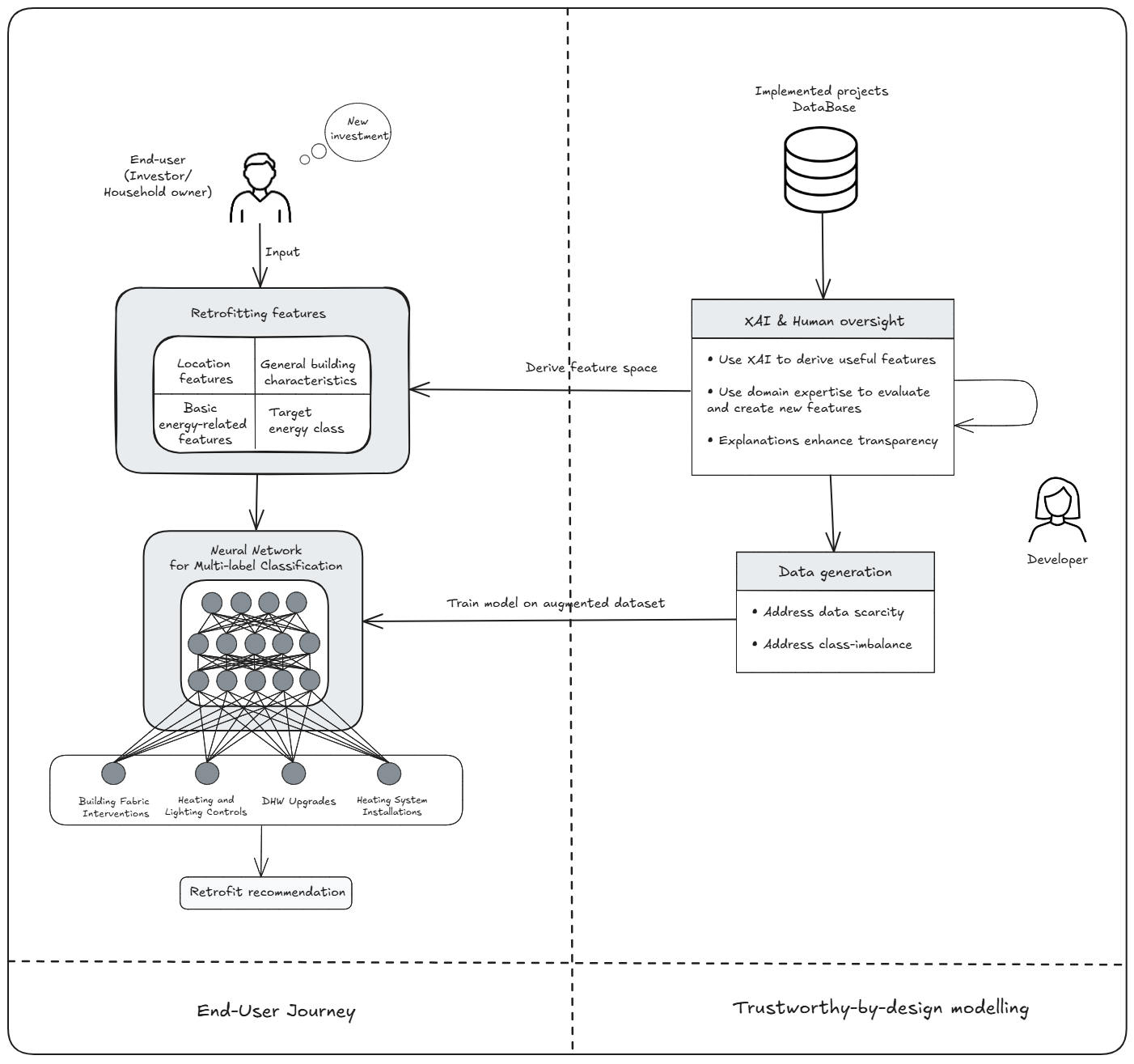}
\caption{Overview of the proposed framework regarding both its design and deployment phase}
\label{fig:methodology}
\end{figure}

This section provides an overview of the decision support framework which is designed to provide high-level energy efficiency retrofit recommendations for building owners and investors. The framework is based on a neural network multi-label classification system developed upon trustworthiness-by-design principles. Regarding the design phase of the framework, an XAI layer is employed to enhance transparency and guide feature engineering in combination with domain knowledge. Additionally, to address data challenges, such as data scarcity and class imbalance, that are common in energy retrofitting datasets, a data augmentation layer is incorporated in the design phase. The proposed framework is summarized in Fig. \ref{fig:methodology}. 

\subsection{Proposed decision support framework for building energy efficiency retrofits}
The problem is formulated as a multi-label classification task, where a neural network classifier directly suggests renovation strategies based on data from previously implemented projects.

The primary goal is to design a user-friendly system in which non-technical end-users, such as household owners, can input details about their building and receive appropriate retrofit recommendations, thereby promoting stakeholder engagement and satisfaction. To this end, the input features are easily accessible and known to users. This minimal feature set includes location-based attributes, general building characteristics, and basic energy-related information that can be derived from EPCs. Additionally, the user specifies the target energy class they wish to achieve, which guides the recommended measures.

The model output comprises the proposed retrofit actions, with a label of 1 indicating that a measure is recommended for implementation to achieve the target and 0 otherwise. Note that more than one measure can be selected for each building, which justifies treating the problem as a multi-label classification task. This approach also facilitates the recommendation of combinations of measures, which is particularly important for supporting deep renovations. To enhance usability, the recommendations are presented under the following categories, making them understandable and aligned with EU retrofit communication practices:
\begin{itemize}
    \item Building Fabric Interventions: Upgrades to the building envelope such as insulation of walls, roof, floors, upgrades of doors and windows etc. 
    \item Heating and Lighting Controls: Enhancements or replacements of various building systems, such as ventilation, lighting and heating control systems to optimize energy use.
    \item Domestic Hot Water (DHW) Upgrades: Improvements to domestic hot water production, storage and distribution systems.
    \item Heating System Installations: Upgrades or replacements of heating systems using renewable energy sources or more efficient technologies to better manage existing fuels
\end{itemize}

\subsection{Classification model}
The proposed methodology employs a multi-layer perceptron (MLP) as the prediction model. The MLP is a deep feed-forward neural network that is widely used in various DL problems due to its simplicity and flexibility \cite{Tzortzis2023TransferSeries}. Its structure enables the straightforward and efficient exploration of a variety of methods and experimental conditions. In general, neural networks have been preferred by researchers in cases involving energy-related data for building analysis, since they are able to handle the complex properties of the data \cite{Mohandes19}. Furthermore, MLPs are inherently capable of predicting multiple targets simultaneously, offering a simple and effective approach for multi-label classification \cite{BOGATINOVSKI2022117215}.

The MLP consists of multiple neurons organized into multiple layers. Each neuron is fully connected to the outputs ($x_i$) of the previous layer's neurons with weights ($w_i$) and has a fixed bias term ($b$). Additionally, an activation function ($f$) is applied to each neuron's output, introducing non-linearity to the model and allowing it to learn complex patterns. Common activation functions include rectified linear unit (ReLU) and sigmoid of Eq. \eqref{eq:sigmoid}. The latter is suitable for our multi-label classification task, since it converts each output into a probability between 0 and 1, indicating the likelihood of each measure being suggested. During the training of the MLP model, the back-propagation algorithm is used to iteratively update the network's weights and biases. The aim of this process is to compute their optimal values that minimize loss function. 

\begin{equation}
Sigmoid\left(x\right) = \frac{1}{1 + e^{-x}}
\label{eq:sigmoid}
\end{equation}

Since the objective of our MLP model is to predict the appropriate energy efficiency measures based on the characteristics of each building, the input layer size of our model is equal to the number of selected features for each building data sample. The output layer size corresponds to the number of classes the model is expected to predict, i.e. the four retrofitting options, while the hidden layers are configured through hyperparameter tuning.

\subsection{Trustworthy System Design}


To address the challenges of data quality and AI trustworthiness, our framework design incorporates two dedicated layers: (i) an explainability layer that enhances transparency, guides feature engineering, and enables human oversight for model developers, and (ii) a data generation layer that leverages advanced generative models to enrich the available datasets and counter class imbalance. Together, these layers form the foundation of a trustworthy-by-design framework that is aligned with both technical reliability and emerging policy requirements, such as those defined in the EU AI Act.

Optionally, the selected dataset features can be adjusted by developers based on data and metadata availability, considering their ease-of-use for the end-user and addressing challenges such as the presence of numerous null values and inconsistent recording formats in the dataset. The features can be further enriched with product features informed by domain knowledge to enhance the value of the available data and uncover hidden patterns, without interfering with user experience as they could be automatically calculated during inference. The XAI layer can be used by the model developer to ensure that transformations contribute meaningfully to model performance.

\subsubsection{Explainability}
The proposed framework integrates an XAI layer to address two key objectives. First, the insights from the explainability analysis are used to interpret the model outputs, clarifying why each retrofit measure is recommended and increasing the reliability of the suggestions. Second, the analysis guides the feature engineering process by identifying features that contribute minimally or unexpectedly to predictions based on SHAP values, informing their removal or modification. It also highlights potential gaps where additional features could enhance model performance and support feature space optimization. Domain knowledge and human oversight are incorporated to critically evaluate the results and guide implementation choices. This process is conducted in iterative XAI rounds, enabling continuous refinement of both model transparency and predictive performance.

As for the explainability approach, we employ SHAP (SHapley Additive ExPlanations), one of the most well-established XAI methods \cite{Lundberg2017}. It is based on the Shapley values from game theory, which reflect the contribution of each feature to the difference between the actual prediction and the average prediction \cite{molnar2022}. The Shapley value for a feature $i$ is calculated as the weighted sum of $i$'s contribution across all possible combinations of feature values using the Eq. \eqref{eq:shapley}:
\begin{equation}
\phi_i = \sum_{S \subseteq F \setminus \{i\}} \frac{|S|!(|F| - |S| - 1)!}{|F|!} \left[ f_{S \cup \{i\}}(x_{S \cup \{i\}}) - f_S(x_S) \right]
\label{eq:shapley}
\end{equation}
where $S$ is any possible subset of features that does not include $i$, $|F|$ is the total number of features, $f_{S \cup \{i\}}(x_{S \cup \{i\}})$ is the model's prediction using only the features in $S$ and $f_S(x_S)$ is the model prediction when $i$ is not included.

The contribution of each feature is quantified using SHAP values, which serve as a measure of feature importance. A larger absolute SHAP value indicates a greater influence of the corresponding feature on the prediction and thus it is relatively more important. Positive SHAP values contribute positively leading to higher output values, while negative SHAP values contribute negatively decreasing the output. In the context of classification problems, this corresponds to the classes with labels 1 and 0 respectively. For more details, readers are encouraged to refer to the original SHAP paper \cite{Lundberg2017}.

Summary plots are used to visualize global explanations, which are helpful to understand the overall behavior of the model. The vertical axis of these plots displays the features ordered by the mean absolute value of their SHAP values, which expresses the feature importance. Hence, the most important feature, i.e. with the highest SHAP value, appears at the top of the plot. The horizontal axis shows the SHAP values and specifically each dot represents an individual data sample and its SHAP value for a specific feature. Finally, the color of each dot reflects the feature value for the specific sample. Shades of blue indicate low values relative to other samples, while shades of red indicate high relative values.

Local explanations, which refer to individual model's predictions, are illustrated through waterfall plots \cite{shap}. These diagrams capture the cumulative property of SHAP values beginning at the bottom of the plot with the base value $E[f(x)]$: the expected prediction of the model if no features were known. Then each line corresponds to a feature and the (normalized) value it receives. It shows how its positive (red) or negative (blue) impact (SHAP value) leads to the final prediction of the sample $f(x)$, displayed at the top of the diagram.

\subsubsection{Data generation}
To address the common challenges of energy retrofitting datasets--such as data scarcity, and class imbalance--we incorporate a data generation layer into our framework. Regarding the method used for data generation, we opted for Conditional Tabular Generative Adversarial Network (CTGAN) \cite{ctgan}, because it offers a streamlined method tailored to tabular data and conditional generation. Specifically, CTGAN is a variant of GANs that was proposed to model tabular data and overcome challenges posed by imbalanced datasets. CTGAN incorporates the mode-specific normalization, which converts continuous features into a vector representation suitable for models like MLPs. This approach deals with the non-Gaussian and multimodal distributions commonly present in tabular data. As regards the class-imbalance in discrete columns, it is handled by a conditional generator and the training-by-sample strategy. By setting specific conditions as input, the data generation process focuses on underrepresented classes, ensuring also that they are involved equally in the training. 

Once the synthetic data is generated, the next crucial step includes its evaluation to verify that it retains properties similar to initial data. It is essential to ensure that the data structure remains consistent and to assess the similarity of column distributions and their relationships between the synthetic and real datasets.

\subsection{Experimental setup} \label{exp_setup}
The experimental setup adopted for the application of the framework was designed to systematically evaluate the performance of the proposed framework under realistic conditions. The MLP model was implemented using PyTorch \cite{pytorch} and specifically PyTorch Lightning \cite{Falcon_PyTorch_Lightning_2019}. With respect to the activation function, we applied ReLU in the hidden layers and sigmoid in the output layer. As regards the optimizer, we selected Adam (Adaptive moment estimation) optimizer which is widely used in related cases.  For the loss function, binary cross-entropy (BCE) loss was chosen. 

The dataset was split into three subsets for the training, optimization and evaluation of the model. Specifically, it was divided with 75\% of the data samples allocated to train/validation set and 25\% to test set. The former was further split into 75\% for training and 25\% for validation. Numerical features were normalized using min-max scaler, while categorical columns were encoded. We also applied early stopping based on the validation loss to prevent overfitting and reduce the computational cost. 

Regarding the hyperparameter tuning, it was implemented by Optuna \cite{Optuna} optimization library in Python. Tree-structured Parzen Estimator (TPE) was used as the sampling strategy in order to select the most promising combinations of hyperparameter values. Moreover, the pruning strategy Median Pruner terminated ineffective training early based on the validation loss. The search space of each hyperparameter is presented in Table \ref{tab:hyperparameters}. We executed 50 trials to select the optimal architecture of the model, learning rate and batch size for training process. 
\begin{table}[h]
    \centering
    \renewcommand{\arraystretch}{1.2} 
    \begin{tabular}{cc}
        \toprule
        \textbf{Hyperparameter} & \textbf{Search space} \\ \midrule
        Number of layers & 2, 3, 4, 5, 6 \\ 
        Layer sizes & 32, 64, 128, 256, 512 \\
        Learning rate & 0.0001, 0.001, 0.01 \\ 
        Batch size & 16, 32, 64, 128 \\ \bottomrule
    \end{tabular}
    \caption{The hyperparameters optimized for MLP and their search spaces}
    \label{tab:hyperparameters}
\end{table}

Model performance was assessed through various metrics commonly used in classification problems. Accuracy is the ratio of correct predictions to total predictions as shown in Eq. \eqref{eq:accuracy}. Precision shows the ratio of correct positive predictions to total positive predictions and it is defined by Eq. \eqref{eq:precision}. Recall is the ratio of correct positive predictions to total actual positive samples as shown in Eq. \eqref{eq:recall}. F1 score refers to the harmonic mean of Precision and Recall according to Eq. \eqref{eq:f1}. In our model we adapted the above metrics to the multi-label classification, i.e. Multilabel Accuracy. They were calculated independently for each label and then the unweighted average was extracted. 

\begin{equation}
Accuracy = \frac{TP + TN}{TP + TN + FP + FN}
\label{eq:accuracy}
\end{equation}

\begin{equation}
Precision = \frac{TP}{TP + FP} 
\label{eq:precision}
\end{equation}

\begin{equation}
Recall = \frac{TP}{TP + FN} 
\label{eq:recall}
\end{equation}

\begin{equation}
F1~score = \frac{2 \times Precision \times Recall}{Precision + Recall}
\label{eq:f1}
\end{equation}

For the explainability analysis, we utilized the SHAP Python library to calculate SHAP values for all target classes. CTGAN was implemented with the Synthetic Data Vault (SDV) \cite{SDV}, a Python library which includes tools for tabular data generation. The hyperparameters were set to the default settings of SDV's CTGAN synthesizer and training was conducted for 800 epochs. The data generation was performed with conditional sampling to address class-imbalance by directing the process toward minority labels. Given the interdependencies between the four labels (energy efficiency measures) due to the multi-label nature of the task, we generated synthetic samples under different conditions to achieve a balanced class distribution for each label.

For the synthetic data evaluation, we used Python library SDMetrics (Synthetic Data Metrics) \cite{sdmetrics}, which is developed to compare synthetic against real tabular data. 
We applied a diagnostic report using SDMetrics, which checks that all columns are included and their data types are preserved (i.e. the data validity) \cite{sdmetrics}. To evaluate the similarity of distributions and correlations of the features, we utilized the SDMetrics quality report, which contains the following metrics serving our purpose effectively \cite{sdmetrics}:
\begin{itemize}
    \item Column Shapes Score \cite{sdmetrics}: It evaluates the overall distribution of each individual column (its similarity across original and synthetic data). KSComplement metric is computed for numerical columns using the Kolmogorov-Smirnov statistical test to measure the distance between distributions. TVComplement is calculated for categorical columns. This metric is derived from the Total Variation Distance between synthetic and real columns, expressing the frequency of each category as a probability. 

    \item Column Pair Trends Score \cite{sdmetrics}: It assesses whether the synthetic data captures relationships between columns. As regards numerical features, Correlation similarity is computed and specifically Pearson correlation coefficient. For pairs of features consisting of two categorical features or one categorical and one numerical, the Contingency similarity metric is used, which compares contingency tables through Total Variation Distance. Regarding the mixed pairs, numerical columns are discretized into bins in order to evaluate the frequency distributions of value combinations.
\end{itemize}

A higher score indicates greater similarity between synthetic and real data. The final results included both the average scores across all columns or column pairs, as well as the individual scores for each column.

The generated data was utilized to train and evaluate the MLP model, following the same experimental setup as applied to the original dataset. The test set remained identical to that used for the initial data to ensure consistency in evaluation. For training, we employed a combination of synthetic data and the remaining original data (excluding the test set). This combined dataset was then partitioned into training and validation subsets, with 75\% allocated for training and 25\% for validation.

\section{Case study I} \label{case1}
\subsection{Data set and feature engineering}
Our first case study applied the proposed framework to a well-established, large public dataset: the English Energy Performance of Buildings Data for England and Wales \cite{englishEPCs}, a publicly available EPC database managed by the UK Department for Levelling Up, Housing and Communities. Specifically, we used the 08/09/2025 version of the dataset, focusing on residential buildings. This dataset contains a wide range of attributes describing building characteristics both before and after the implementation of energy efficiency retrofit measures. Following the guidelines outlined in the methodology regarding feature selection, we concluded to the initial feature set of Table \ref{tab:selected_features_uk}. With respect to the output retrofit measures, the recorded improvements in the dataset were mapped onto the four retrofit categories addressed by our framework (see \ref{appendix_mapping} for further details).

\begin{table}[!h]
\centering
\begin{tabular}{cccc}
\toprule
\textbf{Feature} & \textbf{Description} & \textbf{Unit} & \textbf{Type} \\ \midrule
Current energy rating & Energy class before retrofit & - & Categorical \\
Potential energy rating & Target energy class after retrofit & - & Categorical \\
Property type & Type of property (e.g., house, flat) & - & Categorical \\
County & County where the building is located & - & Categorical \\
Energy consumption current & Annual total energy consumption before retrofit & kWh/m\(^2\) & Numerical \\
Total floor area & Total useful floor area of the building & m\(^2\) & Numerical \\
Mains gas flag & Presence of mains gas (presence of a gas meter & - & Boolean \\
 & or a gas-burning appliance in the dwelling) & & \\
Floor level & Floor level of the property & - & Numerical \\
Flat top storey & Whether the flat is on the top storey & - & Boolean \\
Flat storey count & Number of storeys in the building & - & Numerical \\
Extension count & Number of extensions added to the property & - & Numerical \\
Number habitable rooms & Number of habitable rooms & - & Numerical \\
Number heated rooms & Number of heated rooms & - & Numerical \\
Number open fireplaces & Number of open fireplaces & - & Numerical \\
Wind turbine count & Number of wind turbines & - & Numerical \\
Floor height & Average height of the storey & m & Numerical \\
Photo supply & Percentage of photovoltaic area as a & \% & Numerical \\
 & percentage of total roof area & & \\
Solar water heating flag & Presence of solar water heating & - & Boolean \\
Local authority label & Local authority name & - & Categorical \\
Constituency label & Parliamentary constituency name & - & Categorical \\
Construction age band & Age band when building constructed & - & Categorical \\ \bottomrule
\end{tabular}
\caption{Selected input features for the UK dataset}
\label{tab:selected_features_uk}
\end{table}

We applied data pre-processing to remove null and invalid values. Entries in which the energy class did not clearly improve after the retrofit actions were also removed to ensure unambiguous class upgrades. Furthermore, a harmonization step was performed to ensure consistency in value representation—for example, the ``Floor level'' feature was converted into numerical integers. The final processed dataset comprised 183,541 samples, each corresponding to a distinct building. 

\begin{figure}[htbp]
\centering
\includegraphics[scale=0.5]{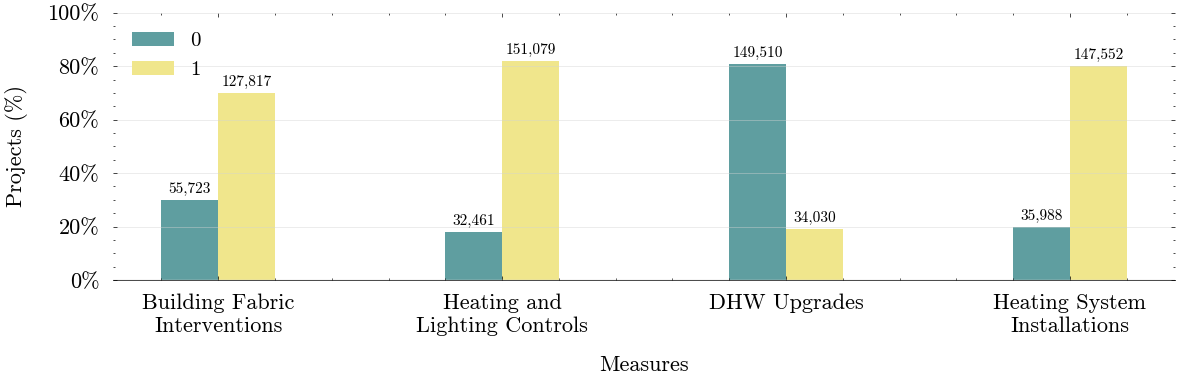}
\caption{Energy efficiency measures (target columns) distributions in initial UK data}
\label{fig:targets_initial_uk}
\end{figure}

Despite its extensive size, the dataset exhibits a notable class imbalance across the energy efficiency improvement measures (outputs), as illustrated in Fig. \ref{fig:targets_initial_uk}. This characteristic makes it particularly suitable for testing our framework on large datasets where class imbalance represents the main limitation.

Note here that through our feature engineering approach we ended up with no additional features for this case study as domain expertise did not reveal useful product features that could be derived from the existing inputs without further complicating the requirements. Nevertheless, this step is less critical in this case, as the large dataset size facilitates the model’s ability to learn patterns effectively without explicit feature engineering.

\subsection{Explainability}

\begin{figure}[h!]
    \centering
    \begin{subfigure}[b]{0.495\textwidth}
        \centering
        \includegraphics[width=\linewidth]{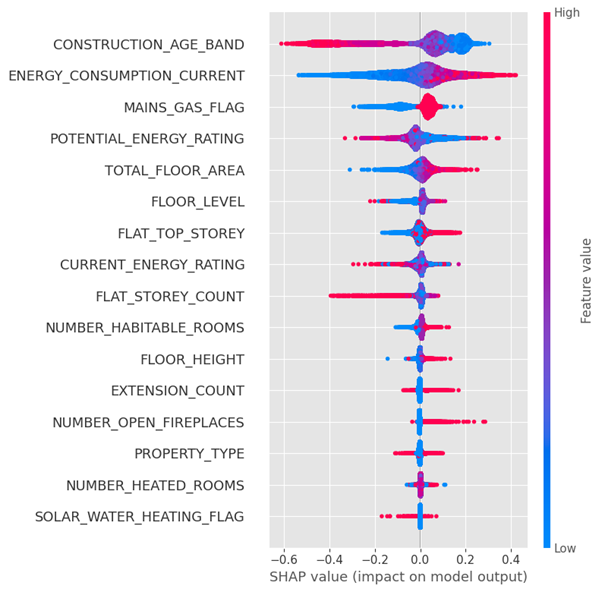}
        \caption{Building Fabric Interventions}
        \label{fig:shap_uk_final_1}
    \end{subfigure}
    \hfill
    \begin{subfigure}[b]{0.495\textwidth}
        \centering
        \includegraphics[width=\linewidth]{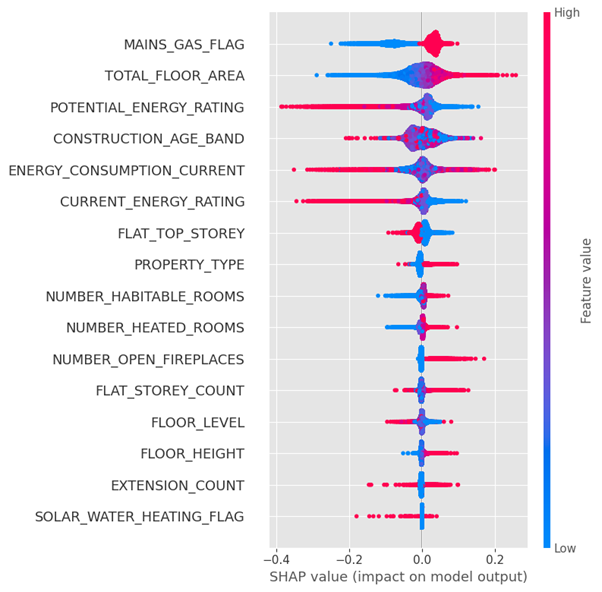}
        \caption{Heating and Lighting Controls}
        \label{fig:shap_uk_final_2}
    \end{subfigure}
    \vspace{0.1cm}
    \begin{subfigure}[b]{0.495\textwidth}
        \centering
        \includegraphics[width=\linewidth]{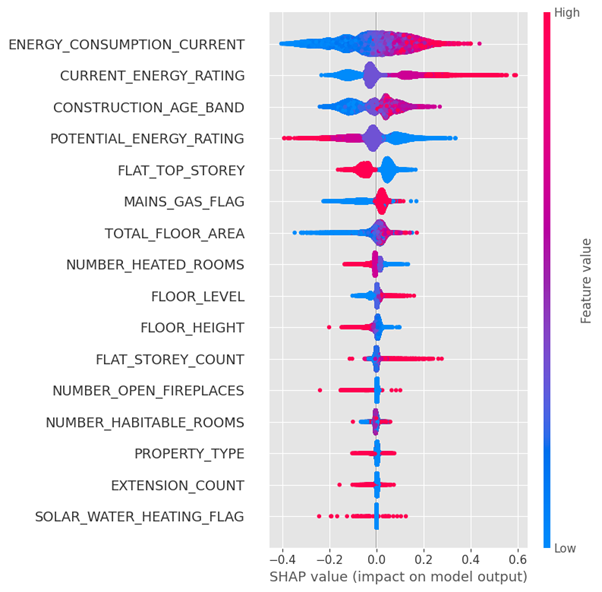}
        \caption{DHW Upgrades}
        \label{fig:shap_uk_final_3}
    \end{subfigure}
    \hfill
    \begin{subfigure}[b]{0.495\textwidth}
        \centering
        \includegraphics[width=\linewidth]{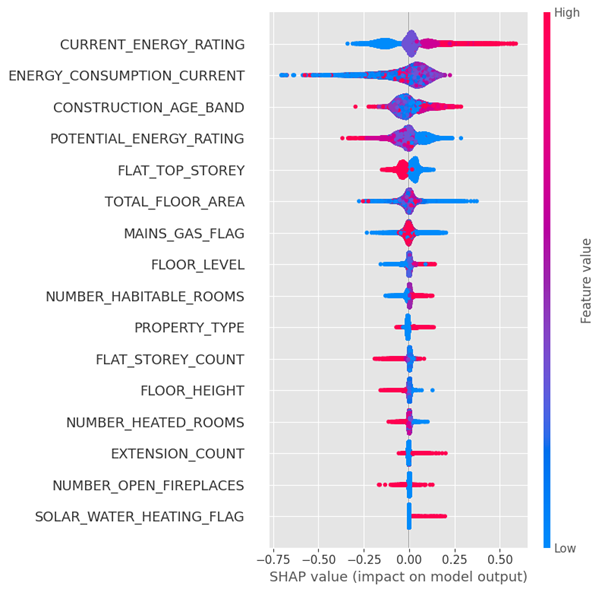}
        \caption{Heating System Installation}
        \label{fig:shap_uk_final_4}
    \end{subfigure}
    \caption{Summary plots of SHAP values for each retrofit class in the UK dataset with the final feature set}
    \label{fig:summary_final_UK}
\end{figure}

 
Examining potentially redundant or uninformative features in the summary plots of the global explanations, it was evident that ``Wind turbine count" and ``Photo supply" consistently appeared among the lowest-ranked features, with very small SHAP values. For ``Wind turbine count", this observation was reinforced by the fact that a large percentage of buildings did not have wind turbines, providing limited information for meaningful conclusions. Furthermore, the location-related features—namely ``Local authority label", ``County", and ``Constituency label"—tended to rank in the upper half of feature importance, but their SHAP values did not exhibit clear patterns. Considering the large number of possible values for these features, this suggested that they could possibly be safely excluded. More details can be found in the \ref{appendix1}, with the Fig. \ref{fig:summary_all_UK} serving as visual evidence. We evaluated model performance after removing these features individually and collectively and found that their removal did not degrade performance; in fact, it improved it as shown in Table \ref{tab:uk_all_vs_final}. Consequently, these features were excluded, with the updated summary plots presented in Fig \ref{fig:summary_final_UK}. For each target class (i.e. ``Building Fabric Interventions", ``Heating and Lighting Controls", ``Heating System Installations", and ``DHW Upgrades") we created a separate plot, which illustrates the distributions of SHAP values revealing the influence of each feature in the decision to implement the corresponding measure.

\begin{table}[h!]
    \centering
    \begin{tabular}{cccccc}
        \toprule
        & \textbf{Accuracy} & \textbf{Precision} & \textbf{Recall} & \textbf{F1 score} \\ \midrule
        {Initial feature set} & 0.825 & 0.768 & 0.767 & 0.740  \\ \hline
        {Final feature set} & 0.836 & 0.792 & 0.774 & 0.763 \\ \bottomrule
    \end{tabular}
    \caption{Model evaluation on UK dataset with initial and final feature set}
    \label{tab:uk_all_vs_final}
\end{table}

Across all output classes, features related to energy consumption, building age, and potential energy ratings consistently demonstrate high influence, underscoring their critical role in predicting intervention outcomes. Multiple patterns emerge, indicating that energy-related features consistently drive intervention recommendations, while structural attributes modulate their importance across different intervention types. Overall, the SHAP analysis provides interpretable insights into the internal decision-making processes of the decision support framework, offering valuable guidance for applying its results. Additionally, more granular insights for individual buildings can be obtained through waterfall plots (Fig. \ref{fig:waterfall_plots_uk}), which illustrate how each feature contributes to specific predictions, enhancing the transparency and interpretability of the model’s decisions.

\subsection{Synthetic data evaluation}

\begin{figure}[htbp]
\centering
\includegraphics[scale=0.5]{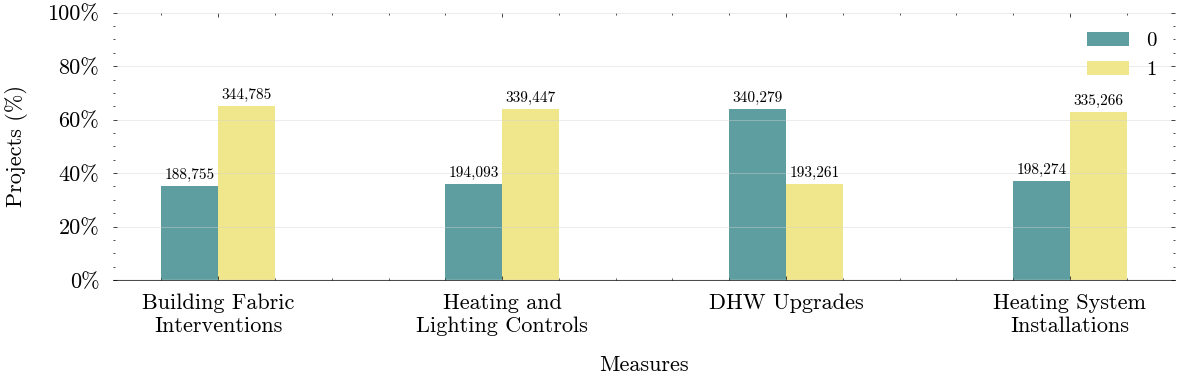}
\caption{Energy efficiency measures (target columns) distributions in synthetic UK data}
\label{fig:targets_synthetic_uk}
\end{figure}

Using CTGAN, we generated 350,000 synthetic samples to balance the distributions of the four output variables, as Fig. \ref{fig:targets_synthetic_uk} illustrates. Before proceeding to the results of the model trained on the synthetic data, we evaluate them:
\begin{itemize}
    \item Diagnostic report: The validity test yielded a result of 100\%, which implies that the basic format and structure of the synthetic data appear identical to those of the real data.
    \item Quality report: We obtain a Column Shapes Score of 88.82\% and a Column Pair Trends Score of 74.56\%, resulting in an average similarity score of 81.69\%.
\end{itemize} 
The evaluation indicates that the synthetic data closely resembles the original dataset.

\begin{figure}[ht]
\centering
\includegraphics[scale=0.5]{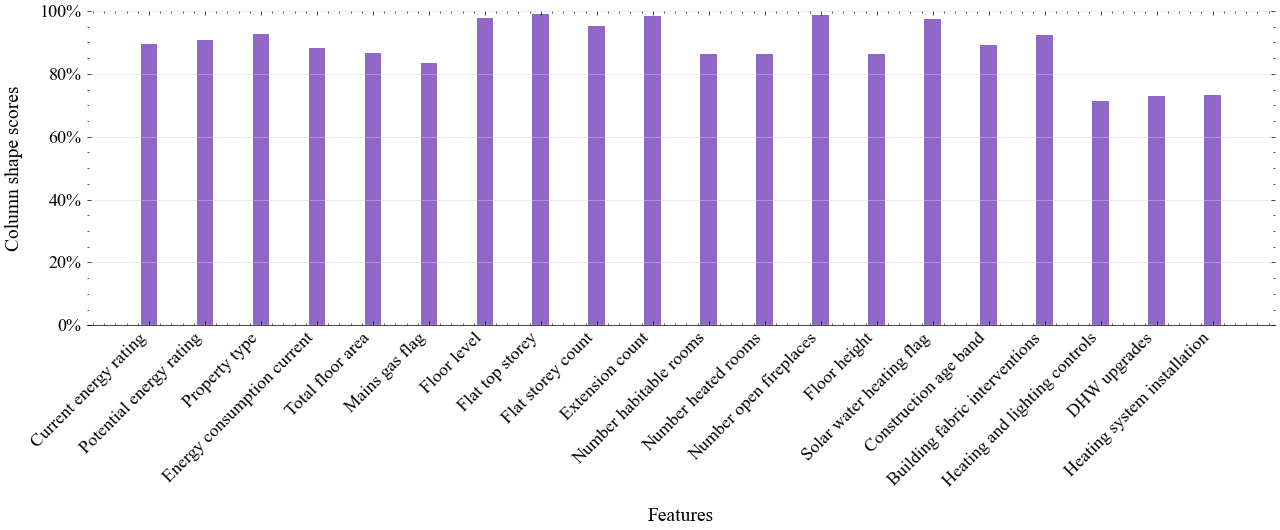}
\caption{Column Shape Score per column of the UK dataset}
\label{fig:column_scores_uk}
\end{figure}

Although the above average scores provide a comprehensive and general evaluation, it is also worth to analyze each column's score separately. Fig. \ref{fig:column_scores_uk} visualizes the Column Shapes Score values that were achieved for each feature. While most features achieved a high score, we observed that the lowest scores corresponded to the four target columns—a reasonable result since their distributions were explicitly defined during the data generation process. Excluding these targets, the average Column Shapes Score increases to 91.64\%, demonstrating a strong overall similarity between the synthetic and real data.

\subsection{Model evaluation on initial and synthetic data}

Table \ref{tab:results_synth_uk} presents the model evaluation results for both the initial and augmented datasets, evaluated on the same real test data.

\begin{table}[h!]
    \centering
    \begin{tabular}{ccccccc}
        \toprule
        \textbf{Train data} & \textbf{Test data} & \textbf{Accuracy} & \textbf{Precision} & \textbf{Recall} & \textbf{F1 score} \\ \midrule
        Initial & Initial & \textbf{0.836} & \textbf{0.792} & 0.774 & 0.763  \\ \hline
        Augmented & Initial & 0.766 & 0.758 & \textbf{0.829} & \textbf{0.785} \\ 
    \bottomrule       
    \end{tabular}
    \caption{Model evaluation on initial and synthetic UK data}
    \label{tab:results_synth_uk}
\end{table}

Initially, the model trained on the dataset containing both synthetic and real data is evaluated on the real test set, exactly as the model trained only on real data allowing a direct comparison. As shown, although the model performs well on the initial dataset, incorporating synthetic samples to augment the training set further improves overall performance, as reflected by the increase in F1 score. The observed drop in accuracy is expected, as this metric is less suitable for evaluating models on class-imbalanced data \cite{sun2009classification}.

\section{Case study II} \label{case2}
\subsection{Data set and feature engineering} \label{data}
Our second case study demonstrates the practical validation and results' reporting of the proposed decision support framework on the ``RETROFIT-LAT" dataset \cite{LEIF_DATA_PAPER} and specifically using its first part named EF\_comp. This data set comprises 198 samples and 80 columns, with each row representing a residential building in Latvia that has implemented energy efficiency measures to improve its energy class under project supported by the Latvian Environmental Investment Fund (LEIF). A complete description of the dataset is provided in \cite{LEIF_DATA_PAPER}.

Based on the methodology requirements for feature selection, we identified an initial feature set, as presented in Table \ref{tab:selected_features}. For the output retrofit targets, we used the  features ``Carrying out construction works'', ``Reconstruction of engineering systems'', ``Water heating system'' and ``Heat installation'', mapping them respectively into the four retrofit categories defined in our study.

\begin{table}[!h]
\centering
\begin{tabular}{cccc}
\toprule
\textbf{Feature} & \textbf{Description}               & \textbf{Unit} & \textbf{Type} \\ \midrule
Region           & Planning region of the building & -             & Categorical   \\ 
The town/village & Town or village of the building         & -             & Categorical   \\ 
County/City      & County or city of the building & -           & Categorical   \\ 
Initial year of exploitation & Year the building was first used & - & Numerical \\ 
Building Total Area & Total area of the building         & m\(^2\)            & Numerical     \\ 
Room volume      & The total volume of rooms in the building                    & m\(^3\)            & Numerical     \\ 
Average floor height & Average height of floors          & m             & Numerical     \\ 
Reference area   & The reference area used in energy  & m²            & Numerical     \\ 
          & performance calculations (Heated area)        &         &    \\ 

Above-ground floors & Number of floors above ground      & -         & Numerical     \\ 
Underground floor & Existence of underground floor        & -         & Boolean     \\ 
Mansard          & Existence of mansard         & -        & Boolean   \\
          & (roof with 4 sloping sides)        &         &    \\ 
Roof floor       & Existence of roof floor   & -        & Boolean   \\
Initial energy class & Energy class before renovation      & -   & Categorical   \\ 
Energy consumption before & Total energy consumption before  & kWh/m² & Numerical \\ 
& renovation & & \\
Energy class after & Energy class after retrofit        & -   & Categorical   \\ \bottomrule
\end{tabular}
\caption{Selected input features for the Latvian dataset}
\label{tab:selected_features}
\end{table}

Data pre-processing was conducted to identify and remove null values and outliers. Null values were minimal, occurring only in ``The Town/village'' feature. Outliers were examined using the z-score method \cite{Kaliyaperumal2015} and our critical observation, which resulted in the removal of one sample.

\begin{figure}[htbp]
\centering
\includegraphics[scale=0.5]{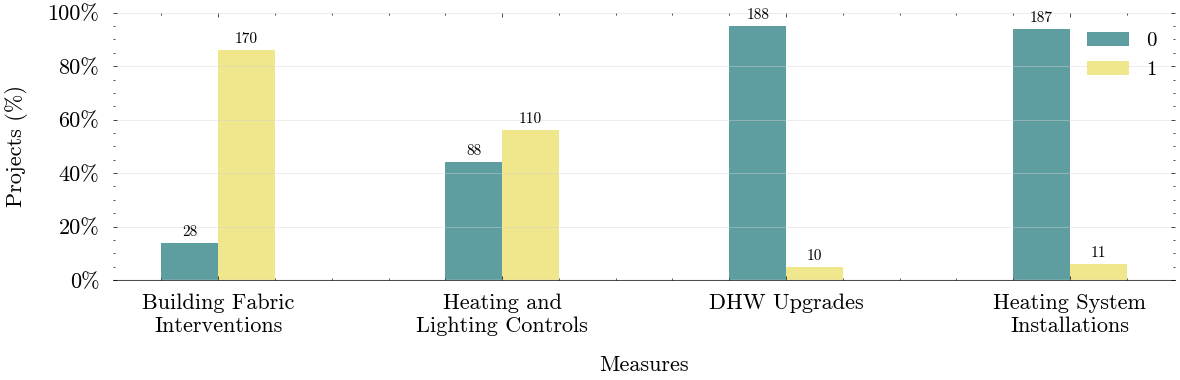}
\caption{Energy efficiency measures (target columns) distributions in initial Latvian data}
\label{fig:targets_initial}
\end{figure}

\begin{figure}[ht]
\centering
\includegraphics[scale=0.6]{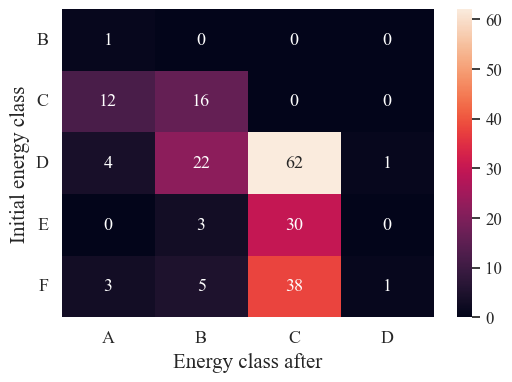}
\caption{Heatmap of \textit{Initial energy class} and \textit{Energy class after} combinations in Latvian dataset}
\label{fig:classes_heatmap}
\end{figure}

Through an exploratory data analysis, we identified several data issues that need to be addressed:

\begin{itemize}
    \item Class-imbalance: The target classes representing energy efficiency measures are highly imbalanced across the dataset. As shown in Fig. \ref{fig:targets_initial}, the percentages of samples with label 1 (implementation) in measures ``Building Fabric Interventions", ``Heating and Lighting Controls", ``DHW Upgrades", ``Heating System Installations" are 86\%, 56\%, 5\% and 6\% respectively. Especially the number of positive samples for the last two measures is extremely small, introducing a bias towards no-implementation and limiting the existence of meaningful patterns. 
    
    \item Data scarcity: An evident data scarcity issue is observed in ``Initial energy class" and ``Energy class after" features. The dataset covers only a limited subset of possible combinations of classes (transitions due to energy efficiency improvements) as illustrated in Fig. \ref{fig:classes_heatmap} with several values being insufficiently represented. For instance, buildings initially in class E have reached class C in their majority, just 3 buildings transitioned to class B and there is no data for transitions into classes D and A. 

    \item Limited size: The dataset contains only 198 data samples, which normally restricts the derived model's ability to generalize, increases the risk of overfitting, and reduces the reliability of predictions.
\end{itemize}

Based on the characteristics of the dataset and the observed scarcity of certain energy class combinations, we introduce a new feature during the feature engineering stage named ``Energy performance delta". The followed approach involves deriving a continuous variable from combinations of other variables, including categorical ones. We speculate that the upgrade of an energy class can be also represented numerically based on regional standards and that this can improve the modelling as certain combinations of initial and final energy classes are rare in the training data. By converting these inputs into continuous variables, the model is expected to capture more nuanced patterns that categorical features alone might miss. This generalization was expected to not only enhance the learning of subtle relationships but also allows the model to make informed predictions on cases that were not explicitly observed during training. In Latvia, energy classes are defined as described in Table \ref{tab:latvian_classes} \cite{Regulations222}, based on energy consumption for heating and the heated area (``Reference area"). Therefore, the new feature ``Energy performance delta" is calculated as in Eq. \ref{eq:new_feature}. 

\begin{table}[h!]
    \centering
    \begin{tabular}{>{\centering\arraybackslash}p{4cm}>{\centering\arraybackslash}p{3cm}>{\centering\arraybackslash}p{3cm}>{\centering\arraybackslash}p{3cm}} 
        \toprule
        \multirow{3}{4cm}{\centering \textbf{Energy efficiency class of building}} 
        & \multicolumn{3}{>{\centering\arraybackslash}p{9cm}}{\centering \textbf{Energy consumption for heating (kWh/m\(^2\)) of residential buildings}} \\ \cline{2-4}
        & \multicolumn{3}{c}{Heated area, m\(^2\)} \\ \cline{2-4}
        & from 50 to 120 & from 120 to 250 & over 250 m\(^2\) \\ \midrule
        A+ & $\leq 35$ & $\leq 35$ & $\leq 30$ \\ 
        A  & $\leq 60$ & $\leq 50$ & $\leq 40$ \\ 
        B  & $\leq 75$ & $\leq 65$ & $\leq 60$ \\ 
        C  & $\leq 95$ & $\leq 90$ & $\leq 80$ \\ 
        D  & $\leq 150$ & $\leq 130$ & $\leq 100$ \\ 
        E  & $\leq 180$ & $\leq 150$ & $\leq 125$ \\ 
        F  & over 180 & over 150 & over 125 \\ \bottomrule
    \end{tabular}
    \caption{Energy classes in Latvia}
    \label{tab:latvian_classes}
\end{table}

\begin{equation}
\Delta E_{\text{class}} = E_{\text{initial}} - E_{\text{final}}
\label{eq:new_feature}
\end{equation}
where \(E_{\text{initial}}\) and \(E_{\text{final}}\) represent the upper limits of the energy consumption ranges for the initial and final energy classes, respectively. A positive value of \(\Delta E_{\text{class}}\) indicates an improvement in energy performance. This feature captures the magnitude of the energy efficiency improvement required for the specified energy class transition, serving as a proxy for the extent of retrofitting needed.

It is worth noting the assumption made in calculating the `Energy performance delta". Since we use the upper limits of each energy class, the resulting value represents the maximum required improvement in energy performance or, in other words, the maximum reduction in energy consumption needed for the transition. While a building may require a smaller change to reach a more efficient energy class, this approach provides a general measure that guarantees the desired upgrade, even in the worst-case scenario. Also, note here that although automated feature-engineering tools exist, this manual feature was intentionally created using domain expertise that is grounded on Latvia’s regulatory definitions for energy efficiency. In this way, we aim to illustrate how designers can use local standards to create meaningful, context-aware features for their own building stock and regulatory environment, boosting both performance and interpretability of their models.

\subsection{Explainability}
\begin{figure}[h!]
    \centering
    \begin{subfigure}[b]{0.49\textwidth}
        \centering
        \includegraphics[width=\linewidth]{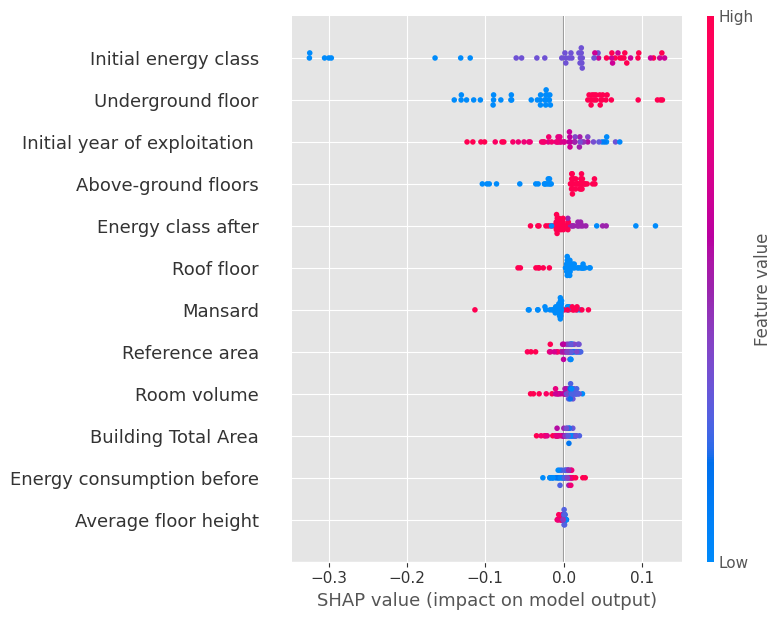}
        \caption{Building Fabric Interventions}
        \label{fig:sum_carr}
    \end{subfigure}
    \hfill
    \begin{subfigure}[b]{0.49\textwidth}
        \centering
        \includegraphics[width=\linewidth]{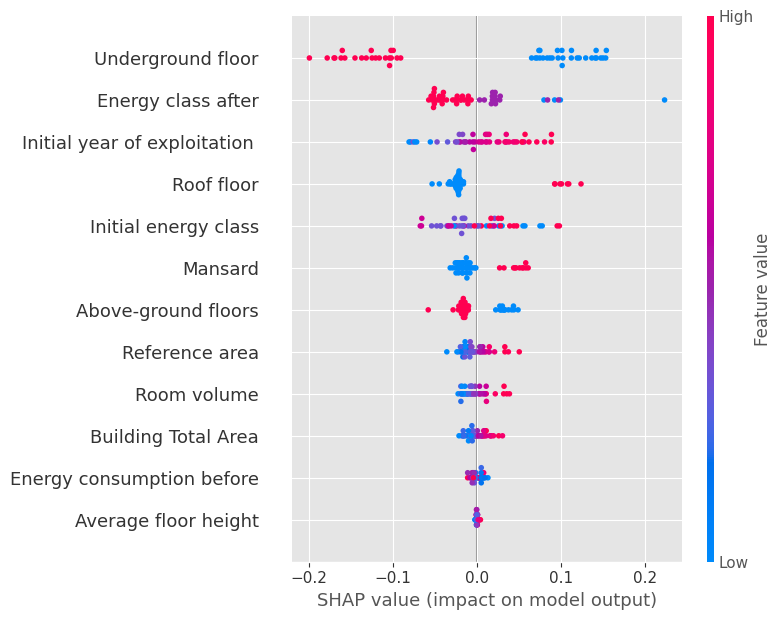}
        \caption{Heating and Lighting Controls}
        \label{fig:sum_rec}
    \end{subfigure}
    \vspace{0.1cm}
    \begin{subfigure}[b]{0.49\textwidth}
        \centering
        \includegraphics[width=\linewidth]{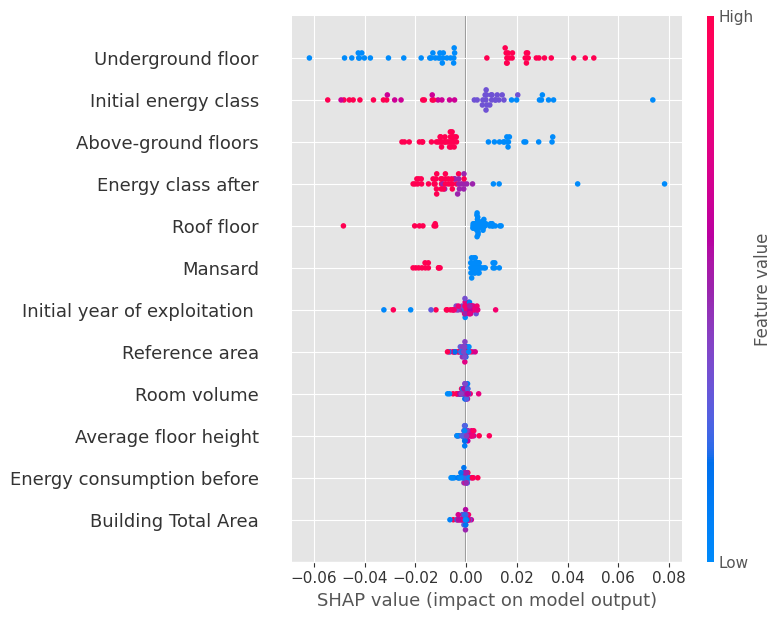}
        \caption{DHW Upgrades}
        \label{fig:sum_wat}
    \end{subfigure}
    \hfill
    \begin{subfigure}[b]{0.49\textwidth}
        \centering
        \includegraphics[width=\linewidth]{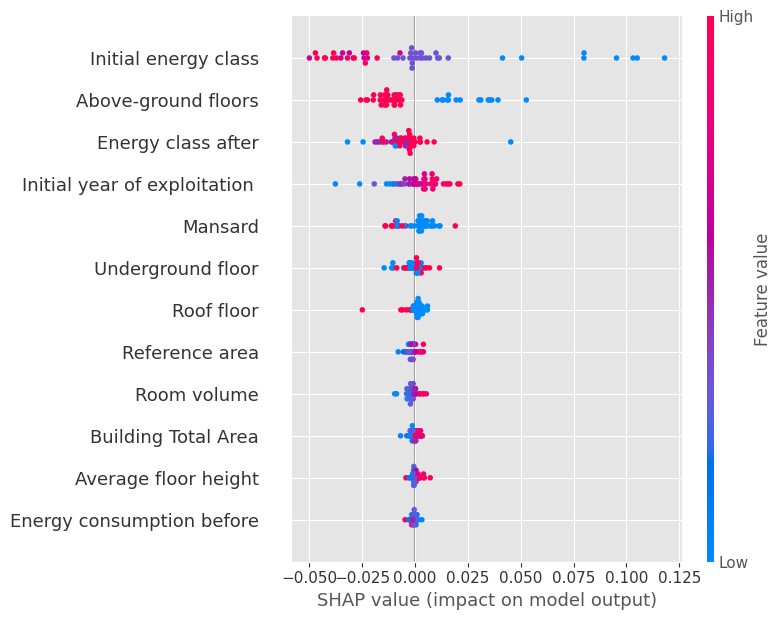}
        \caption{Heating System Installations}
        \label{fig:sum_heat}
    \end{subfigure}
    \caption{Summary plots of SHAP values for each retrofit class in the Latvian dataset}
    \label{fig:sum_plots}
\end{figure}

Using the initial set of features (as defined in Table \ref{tab:selected_features}), it was observed that location-related features --namely ``The town/village", ``County/City" and ``Region"-- had the greatest contribution across all measures. This was quite an unexpected finding that could imply model overfitting to spurious correlations among features and potentially shortcut learning due to limited dataset size \citep{geirhos2020shortcut}. One possible reason stemmed from the fact that these features exhibited large cardinality that resulted in a limited sample count for each unique value (cardinality of 122 and 68 respectively for a total number of 198 samples). Additional information can be found in \ref{appendix2}, where Fig. \ref{fig:sum_plots_loc} serves as visual support. To further reinforce this claim, Table \ref{tab:location_features} lists the results of a model with and without the inclusion of location features as predictors. It is evident that the predictive capacity of ML models is hardly affected by location features, indicating that it is safe to exclude them from our feature set. At the same time, this choice enhances the generalization of the model as these features refer to specific cities and regions of Latvia, hence preventing inference to buildings located at alternative, previously unseen geographical areas. Fig. \ref{fig:sum_plots} depicts the summary plots after the removal of the location-related features. 

\begin{table}[h!]
    \centering
    \begin{tabular}{cccccc}
        \toprule
        & \textbf{Accuracy} & \textbf{Precision} & \textbf{Recall} & \textbf{F1 score} \\ \midrule
        {With location features} & 0.828 & 0.352 & 0.417 & 0.384  \\ \hline
        {Without location features} & 0.834 & 0.358 & 0.402 & 0.375 \\ \bottomrule
    \end{tabular}
    \caption{Model evaluation on Latvian dataset with and without location features: \textit{The town/village, County/City, Region}}
    \label{tab:location_features}
\end{table}

Regarding feature importance, one of the most prominent features appears to be the ``Initial energy class", which ranks at the first place in ``Building Fabric Interventions" and ``Heating System Installations" and at the second in ``DHW Upgrades". For instance, in the first summary plot (Fig. \ref{fig:sum_carr}), higher values of the ``Initial energy class" feature correspond to positive SHAP values (indicating a tendency to predict 1 for the measure), while lower values are associated with negative SHAP values (indicating a tendency to predict 0). The encoding of this categorical feature was performed using a label encoder, preserving the class order as follows: A:1, B:2, C:3, D:4, E:5, F:6. This implies that the measure ``Building Fabric Interventions" is more likely to be applied to buildings with the lowest energy classes. This result is quite intuitive, as low energy efficiency typically requires improvements to the building envelope and insulation, which are critical determinants of overall energy performance.

Subsequently, the``Underground floor" feature also stands out as a significant factor. The latter is binary with value 1 for buildings with a basement and 0 for those without. From Fig. \ref{fig:sum_carr} it can be inferred that buildings with an underground floor (high values - red dots, i.e., 1) are more likely to adopt the measure, whereas the opposite holds for buildings without underground floors (low values - blue dots, i.e., 0). While a similar trend is observed for the ``DHW Upgrades" measure, the opposite results are derived for the ``Heating and Lighting Controls". For ``Heating System Installations", however, the existence of an underground floor presents mixed results. Similarly, ``Above-ground floors" emerges as an impactful feature, showing a clear separation of its values. Buildings with more floors are more likely to carry out construction works, while fewer floors are associated with the rest of measures. 

Similar conclusions can be drawn for other combinations of features and target classes. Overall, it is evident that both the initial and final energy classes are pivotal in the selection of retrofit measures, as anticipated. Additional significant features include those related to the number of floors, the year of construction, and the two roofing types. The remaining features also contribute to the predictions although their impact is generally represented by lower SHAP values.

As an additional observation, the SHAP values for ``Heating System Installations" and ``DHW Upgrades" are generally lower compared to the other two measures. This can be attributed to their smaller number of positive samples in the dataset. It is worth noting that the results are influenced by class-imbalance and other data issues, which affect prediction performance.

To analyze the exact contributions of each feature to the model's specific recommendations we proceed with SHAP's local explanations. To illustrate them we use the waterfall plots of each individual prediction with indicative diagrams for one selected data sample in Fig. \ref{fig:waterfall_plots}. By examining the waterfall plots of each individual data point, we obtain an overview of the influence of each feature in terms of both amplitude and direction. Overall, we observe several consistencies with the global average patterns but variations as well. Therefore, no safe conclusions can be drawn, as each observation requires individual inspection, but it serves as a transparency mechanism, providing insights into the model’s reasoning.

We then proceed with validating the creation of the ``Energy performance delta" feature. The first evaluation of this newly created feature is conducted by examining its contribution to model performance. Its introduction has led to improvements in all metrics as shown in Table \ref{tab:new_feat}. Consequently, as the new feature has been proven successful, it will be incorporated as a default feature from now on. Nonetheless, we should note that this feature is derived based on the Latvian energy standards, therefore resulting to limited model generalization. Hence, it introduces a trade-off between performance and generalization.

\begin{table}[h!]
    \centering
    \begin{tabular}{cccccc}
        \toprule    
        & \textbf{Accuracy} & \textbf{Precision} & \textbf{Recall} & \textbf{F1 score} \\ \midrule
        {Without the new feature} & 0.834 & 0.358 & 0.402 & 0.375  \\ \hline
        {With the new feature} & 0.842 & 0.384 & 0.415 & 0.401 \\ \bottomrule
    \end{tabular}
    \caption{Model evaluation on Latvian dataset with and without the new feature \textit{Energy performance delta}}
    \label{tab:new_feat}
\end{table}

By repeating the explainability analysis with the inclusion of the new feature ``Energy performance delta" (Fig. \ref{fig:sum_plots_new}) its contribution is confirmed. More specifically, regarding feature importance, it appears to be in ninth position out of thirteen for ``Building Fabric Interventions" measure, slightly lower for the ``Heating and Lighting Controls" and among the five most important for the other two measures. This can be attributed to the fact that measures with fewer positive samples (``Heating System Installations", ``DHW Upgrades") gain significant advantages from this auxiliary feature, while the ``Heating and Lighting Controls", having sufficient samples, experiences comparatively less impact. A slight redistribution of the other features' importance is also observed. One of the most notable changes is ``Energy class after" which gets more significance in the predictions with its impact becoming clearer. Even in cases where the contribution was previously mixed, such as the ``Building Fabric Interventions" measure, higher feature values now correspond to higher positive SHAP values, and lower feature values to lower negative ones. However, the local explanations still reveal that the contribution of ``Energy performance delta" can vary depending on the individual sample context.

\subsection{Synthetic data evaluation}
The evaluation of the 800 generated samples reported a 100\% Diagnostic Report score and an 78.27\% Quality Report score, consisting of a Column Shapes Score of 78.27\% and a Column Pair Trends Score of 65.57\%. These results indicate that synthetic data closely resemble the original data in terms of individual column distributions, while capturing relationships between features proves more challenging \cite{Mobeen2024}. However, given the limitations of the initial dataset (i.e., limited quantity and class imbalance) used to train the CTGAN, this result is both anticipated and acceptable for the objectives of our study.

\begin{figure}[ht]
\centering
\includegraphics[scale=0.5]{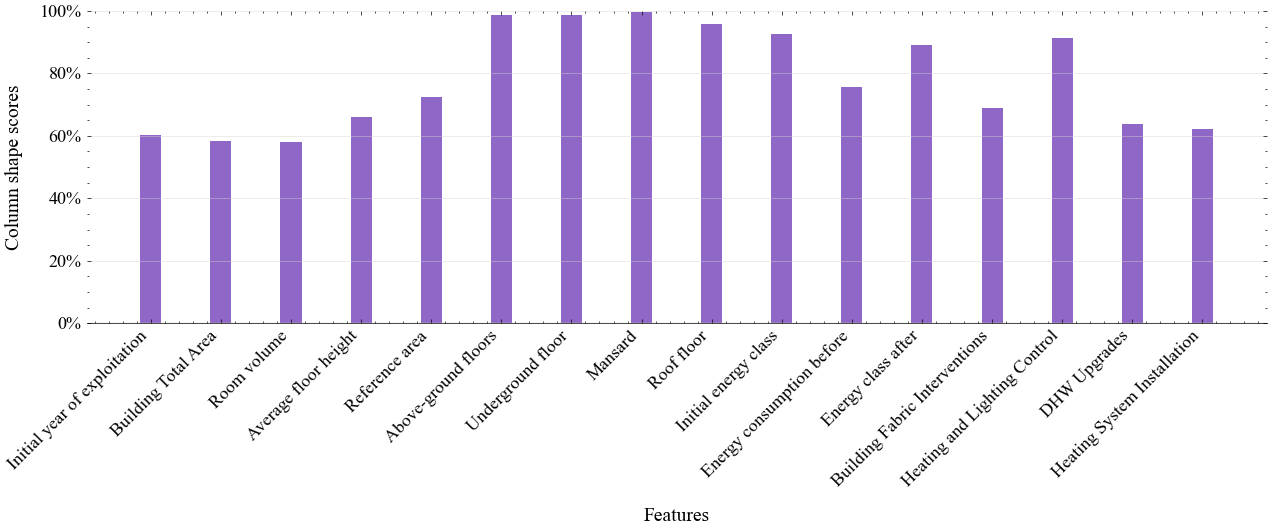}
\caption{Column Shape Score per column of the Latvian dataset}
\label{fig:column_scores}
\end{figure}

When analyzing the Column Shapes Scores for each feature individually in Fig. \ref{fig:column_scores}, a difference in shape quality between categorical and numerical columns is apparent, with the former outperforming the latter. As it is generally found \cite{sdmetrics}, discrete columns (categorical and boolean) with a small number of well-defined categories result in more accurate synthetic data distributions. This happens because they are inherently simpler to model compared to continuous columns, which span large numerical ranges. An exception is observed for the target columns, whose distributions are determined by the conditions applied during data generation process.

\subsection{Model evaluation on initial and synthetic data}
Table \ref{tab:results_synth} shows the model evaluation when trained on the initial and synthetic datasets respectively.

\begin{table}[h!]
    \centering
    \begin{tabular}{ccccccc}
        \toprule
        \textbf{Train data} & \textbf{Test data} & \textbf{Accuracy} & \textbf{Precision} & \textbf{Recall} & \textbf{F1 score} \\ \midrule
        Initial & Initial & \textbf{0.842} & 0.384 & 0.415 & 0.401  \\ \hline
        Augmented & Initial & 0.654 & \textbf{0.408} & \textbf{0.639} & \textbf{0.446} \\ 
     \bottomrule       
    \end{tabular}
    \caption{Model evaluation on initial and synthetic Latvian data}
    \label{tab:results_synth}
\end{table}

Regarding the training on the initial dataset, the precision, recall and F1 score are quite low, confirming the dataset's challenges, already identified in Section \ref{data}, affect the model performance negatively. Conversely, accuracy scores a high value, but it can be misleading, since in class-imbalanced problems like ours this metric mainly focuses on the majority classes \cite{sun2009classification}.

\begin{figure}[htbp]
\centering
\includegraphics[scale=0.5]{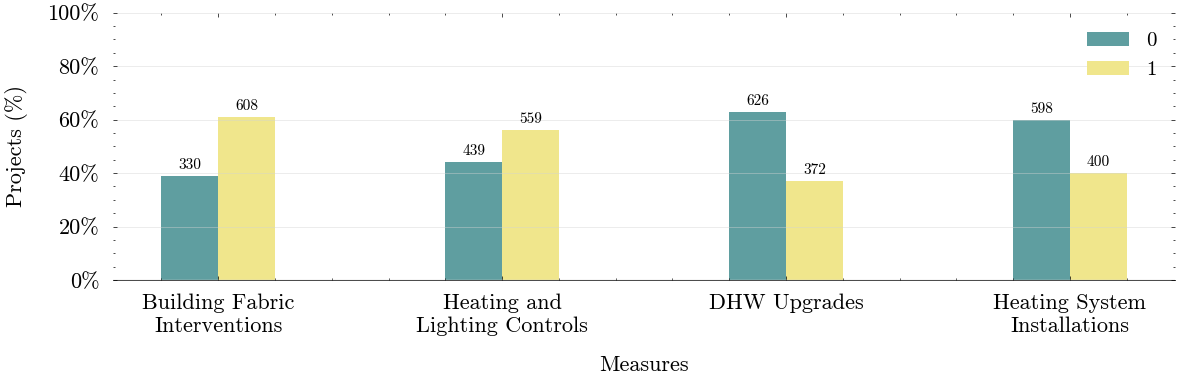}
\caption{Energy efficiency measures (target columns) distributions in synthetic Latvian data}
\label{fig:targets_synth}
\end{figure}

The augmented dataset has mitigated significantly the class-imbalance issue, as illustrated in Fig. \ref{fig:targets_synth} which shows the more balanced target distributions after the data generation process. Although accuracy decreases, this reduction is expected and acceptable since this metric is now more representative and reflects the actual capability of the model. More interestingly, it is obvious that the enrichment benefited the model performance in terms of precision, recall and F1 score. Recall exhibits the greatest improvement from 41.5\% to 63.9\%, indicating a better detection of suitable recommendations. At the same time, the increase in precision and F1 score imply a higher proportion of accurate predicted renovations and an overall enhancement in performance respectively. 

\section{Conclusions and future work} \label{conclusions}

This study proposes a novel trustworthy AI-based decision support framework for recommending energy efficiency retrofit measures in residential buildings, using neural network-based multi-label classification. The framework is directed to non-technical end-users, such as building owners and investors, requiring only a minimal set of inputs derived from basic building characteristics, location information, and EPC-derived energy features. By framing the problem as a multi-label classification task, the framework directly recommends combinations of measures, supporting deep renovations and reducing cognitive load for stakeholders. An additional key contribution is the trustworthy-by-design approach, aligned with emerging regulatory and ethical guidelines. An XAI layer guides feature engineering, ensures model interpretability, and enables human oversight for model developers, enhancing transparency. A GAN-based data augmentation layer mitigates data scarcity and class imbalance, enabling robust performance even under constrained data conditions.

Our framework was validated on two datasets: the larger and mildly imbalanced UK EPC dataset, and the much smaller Latvian RETROFIT-LAT dataset, which exhibits severe data scarcity and class imbalance. The experiments showed that our framework adapted well to contrasting data conditions and can improve recall by up to 53\% --with smaller but consistent gains in precision and F1 score—- compared to a baseline model trained without feature engineering or augmentation. The Latvian case benefited the most, as its limited size and extreme imbalance made the impact of the model more pronounced. In contrast, the UK dataset, with over 30,000 minority-class samples and only one target variable dominated by label 0, started from a higher baseline performance, making imbalance less problematic for identifying positive recommendations that in theoretically correspond to actual realised retrofits. Overall, the results demonstrate that the proposed framework can accommodate data limitations and effectively apply augmentation or feature-engineering strategies where needed, yielding the strongest improvements in small and highly imbalanced datasets.


In summary, the proposed decision support framework offers a practical, interpretable, and generalizable tool for guiding energy retrofit decisions, balancing technical reliability, usability, and regulatory compliance. At the same time it allows for adaptation across different building stocks and regulatory contexts. The framework provides value to i) end-users (building owners, investors) by raising citizen awareness and promoting sustainable retrofit practices; ii) to developers by facilitating trustworthy model design and deployment, and iii) to energy agencies, policymakers, and investors seeking to create similar systems within the EU energy efficiency investment landscape. Its design ensures that stakeholders can leverage data-driven insights to prioritize retrofit investments effectively, even when dealing with challenging datasets. Finally, the practical value of our framework has already been demonstrated through the integration of its core base into AI4EF (Artificial Intelligence for Energy Efficiency) \cite{TZORTZIS2025102172}, validated by a diverse group of stakeholders including policymakers, energy auditors, IT specialists, and researchers, whose feedback ensured that the framework evolved in a user-centric and accessible manner. 

Future directions include expanding to additional regions, incorporating updated renovation outcomes, exploring alternative data generation techniques and model architectures. Our ongoing research focuses on transfer learning and lifelong learning strategies towards expanding the model to new data to extend its scope and impact, while retaining already acquired knowledge.

\section*{Acknowledgments}
This work has been funded by the European Union’s Horizon Europe Research and Innovation programme under the EnergyGuard project, grant agreement No. 101172705. The sole responsibility for the content of this paper lies with the authors; the paper does not necessarily reflect the opinion of the European Commission.

\newpage
\bibliographystyle{elsarticle-num} 
\bibliography{references}

\newpage
\appendix
\section{Additional summary plots of case study I} \label{appendix1}
\begin{figure}[H]
    \centering
    \begin{subfigure}[b]{0.495\textwidth}
        \centering
        \includegraphics[width=\linewidth]{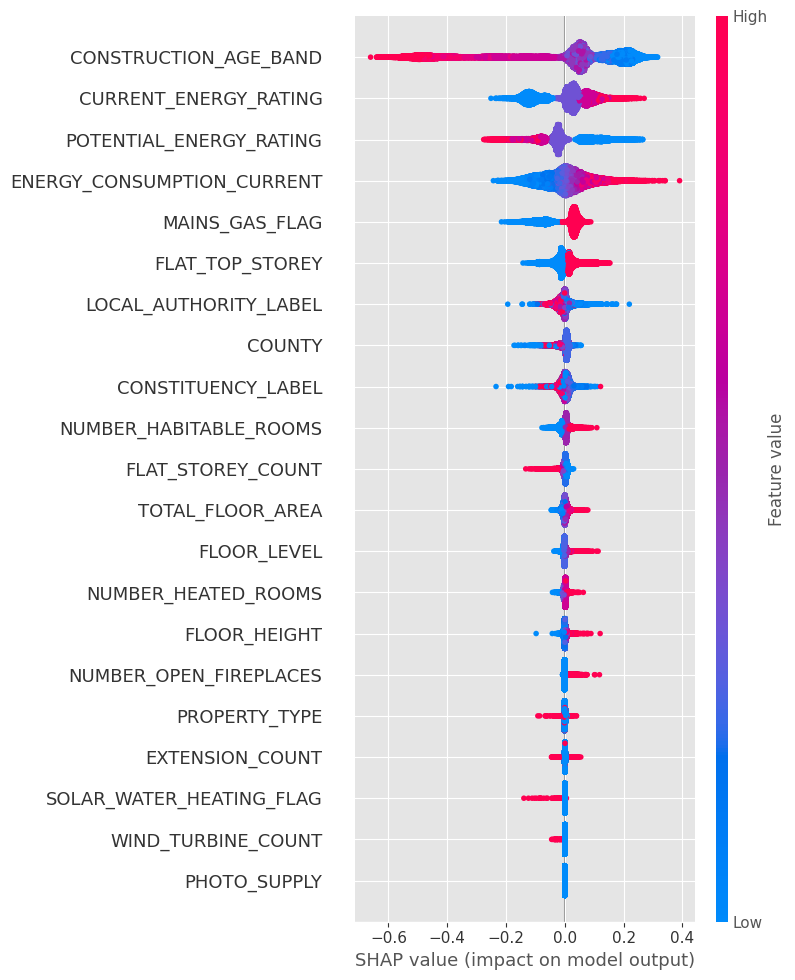}
        \caption{Building Fabric Interventions}
        \label{fig:shap_uk_all_1}
    \end{subfigure}
    \hfill
    \begin{subfigure}[b]{0.495\textwidth}
        \centering
        \includegraphics[width=\linewidth]{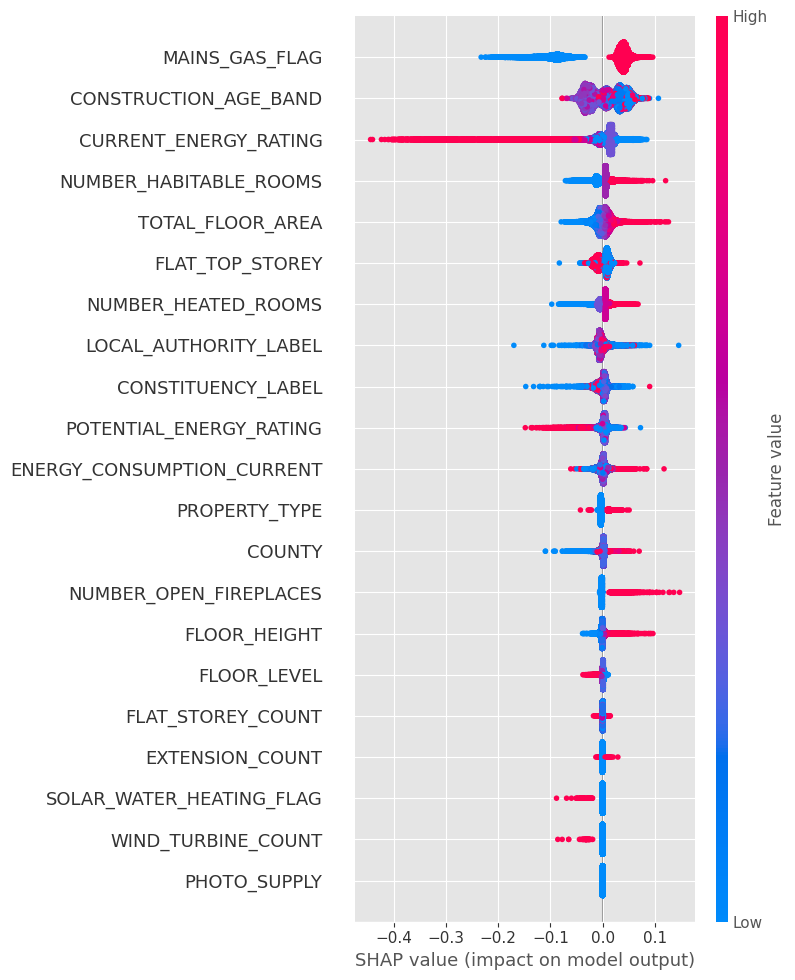}
        \caption{Heating and Lighting Controls}
        \label{fig:shap_uk_all_2}
    \end{subfigure}
    \vspace{0.1cm}
    \begin{subfigure}[b]{0.495\textwidth}
        \centering
        \includegraphics[width=\linewidth]{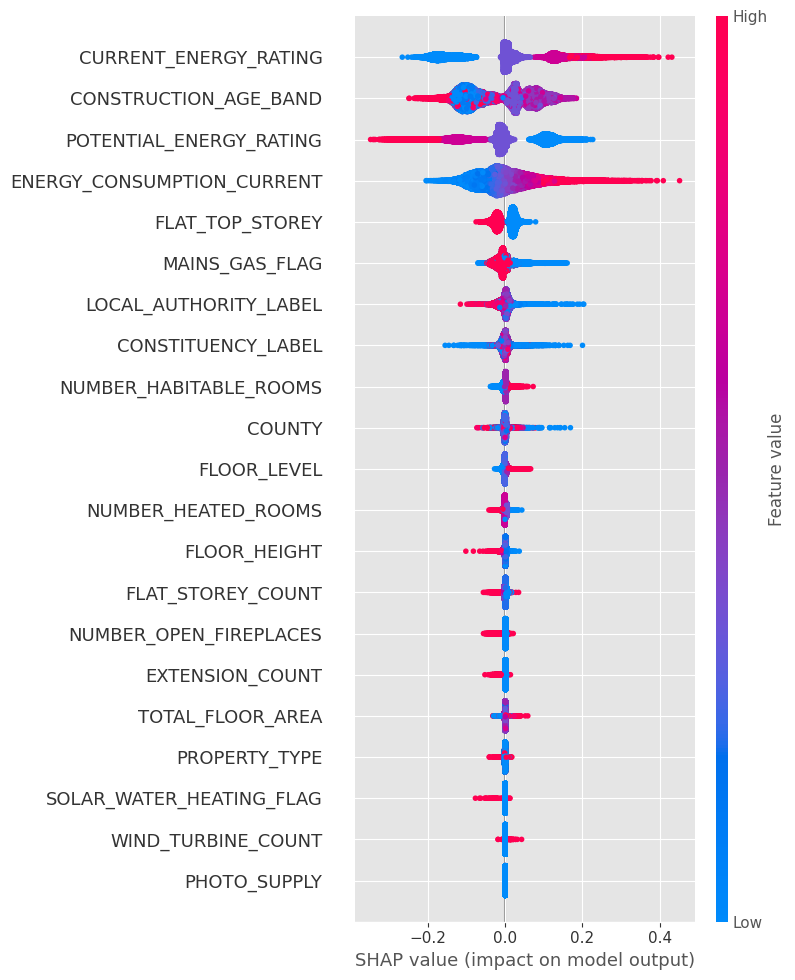}
        \caption{DHW Upgrades}
        \label{fig:shap_uk_all_3}
    \end{subfigure}
    \hfill
    \begin{subfigure}[b]{0.495\textwidth}
        \centering
        \includegraphics[width=\linewidth]{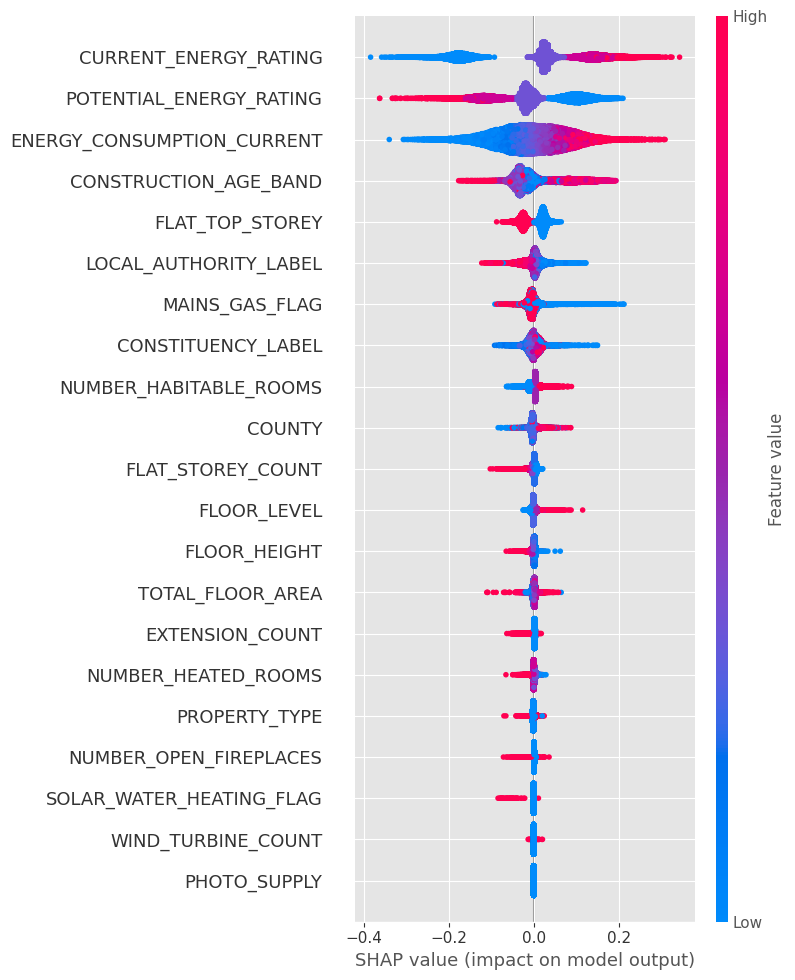}
        \caption{Heating System Installation}
        \label{fig:shap_uk_all_4}
    \end{subfigure}
    \caption{Summary plots of SHAP values for each class with the initial feature set in the UK dataset}
    \label{fig:summary_all_UK}
\end{figure}

\newpage
\section{Additional summary plots of case study II} \label{appendix2}

\begin{figure}[h!]
    \centering
    \begin{subfigure}[b]{0.49\textwidth}
        \centering
        \includegraphics[width=\linewidth]{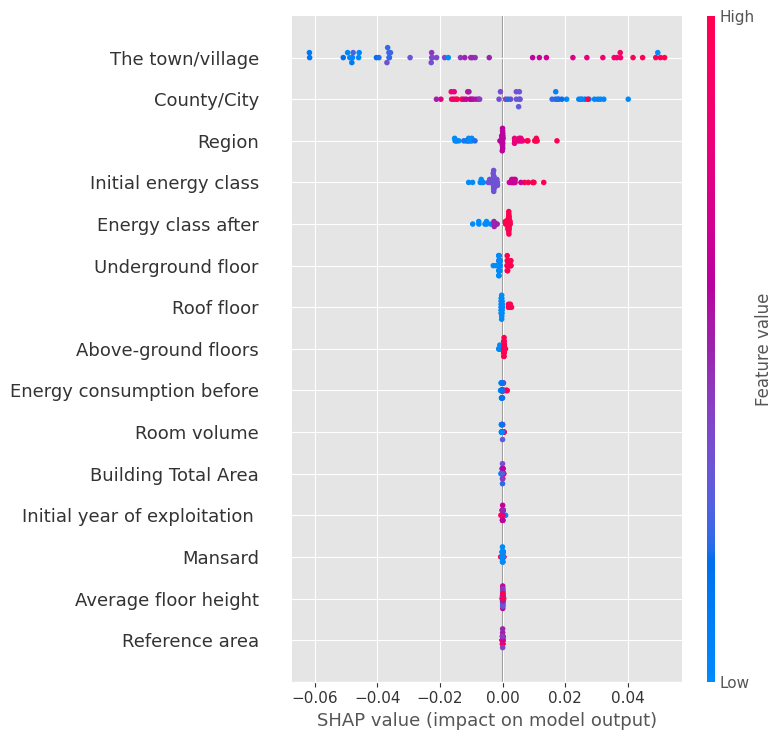}
        \caption{Building Fabric Interventions}
        \label{fig:sum_carr_loc}
    \end{subfigure}
    \hfill
    \begin{subfigure}[b]{0.49\textwidth}
        \centering
        \includegraphics[width=\linewidth]{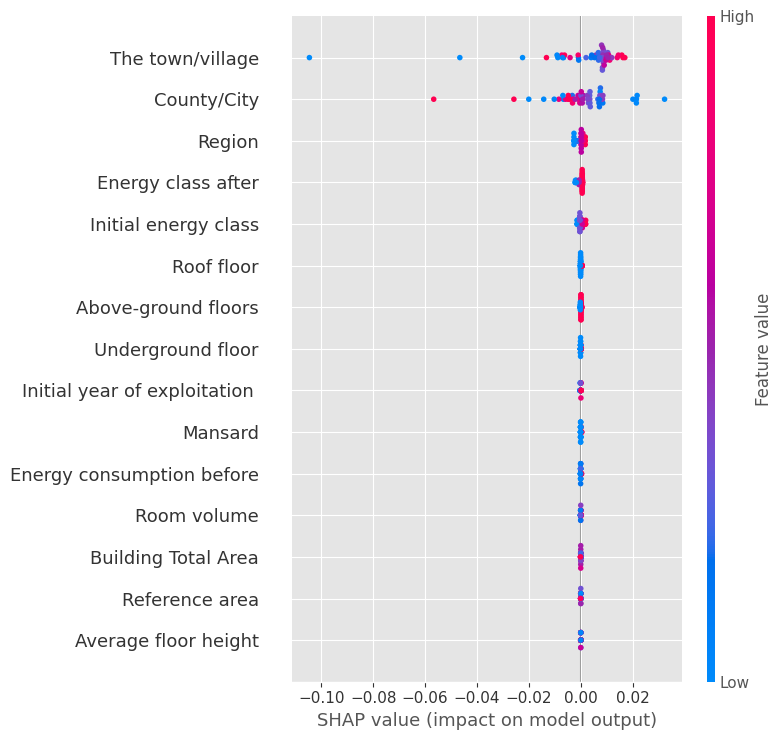}
        \caption{Heating and Lighting Controls}
        \label{fig:sum_rec_loc}
    \end{subfigure}
    \vspace{0.1cm}
    \begin{subfigure}[b]{0.49\textwidth}
        \centering
        \includegraphics[width=\linewidth]{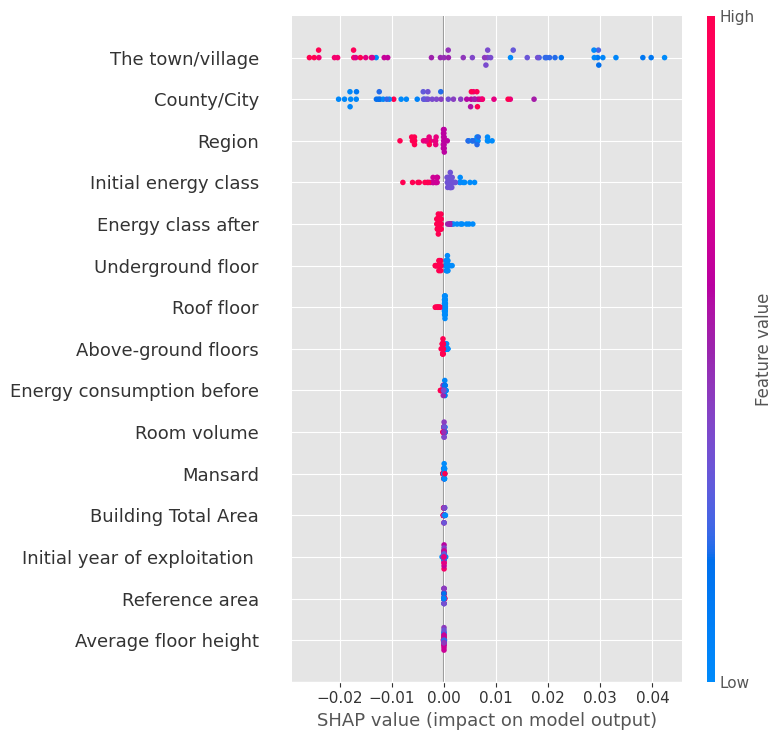}
        \caption{DHW Upgrades}
        \label{fig:sum_wat_loc}
    \end{subfigure}
    \hfill
    \begin{subfigure}[b]{0.49\textwidth}
        \centering
        \includegraphics[width=\linewidth]{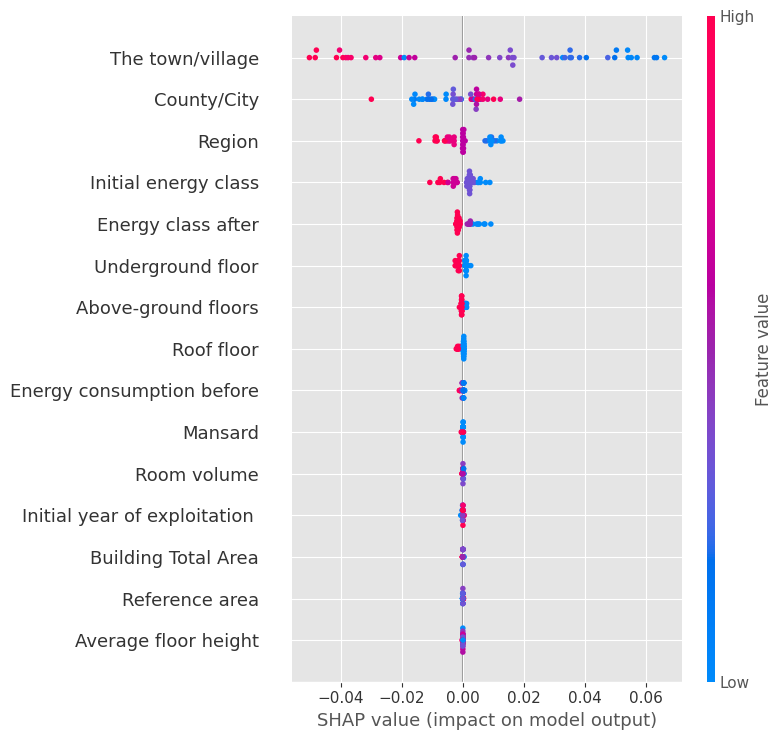}
        \caption{Heating System Installations}
        \label{fig:sum_heat_loc}
    \end{subfigure}
    \caption{Summary plots of SHAP values for each retrofit class including location features in the Latvian dataset}
    \label{fig:sum_plots_loc}
\end{figure}

\newpage

\begin{figure}[h!]
    \centering
    \begin{subfigure}[b]{0.49\textwidth}
        \centering
        \includegraphics[width=\linewidth]{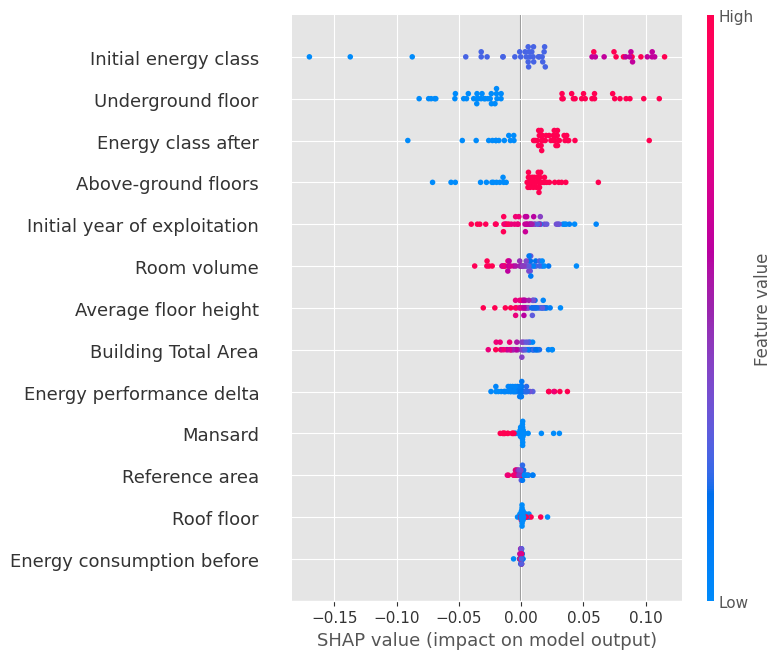}
        \caption{Building Fabric Interventions}
        \label{fig:sum_carr_new}
    \end{subfigure}
    \hfill
    \begin{subfigure}[b]{0.49\textwidth}
        \centering
        \includegraphics[width=\linewidth]{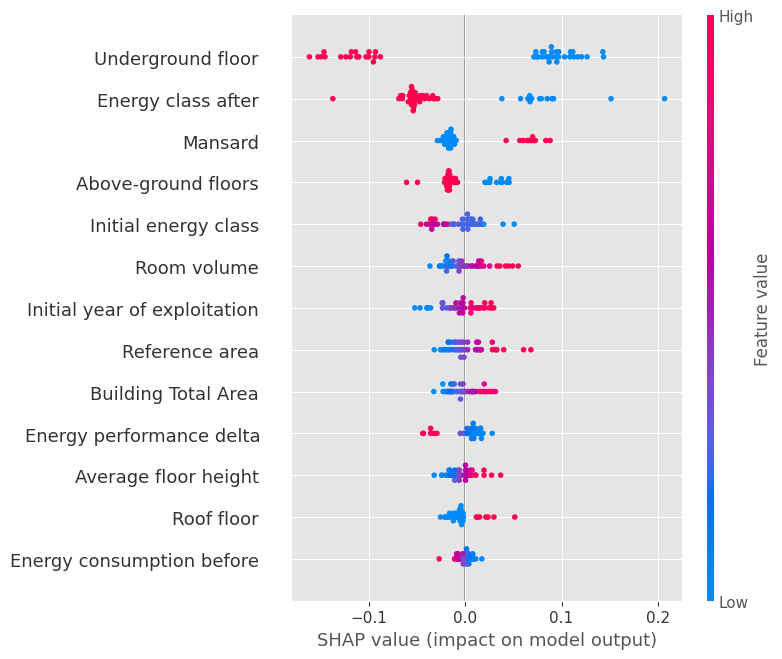}
        \caption{Heating and Lighting Controls}
        \label{fig:sum_rec_new}
    \end{subfigure}
    \vspace{0.1cm}
    \begin{subfigure}[b]{0.49\textwidth}
        \centering
        \includegraphics[width=\linewidth]{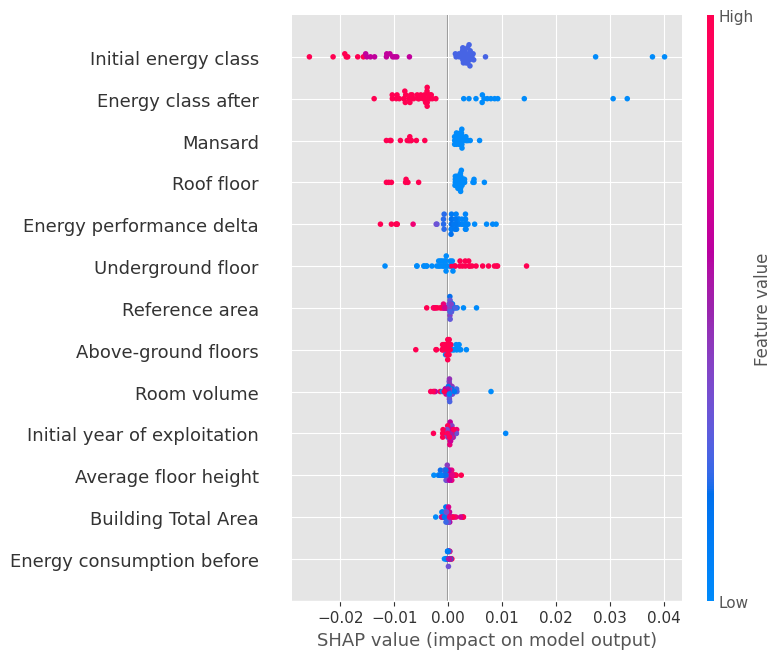}
        \caption{DHW Upgrades}
        \label{fig:sum_wat_new}
    \end{subfigure}
    \hfill
    \begin{subfigure}[b]{0.49\textwidth}
        \centering
        \includegraphics[width=\linewidth]{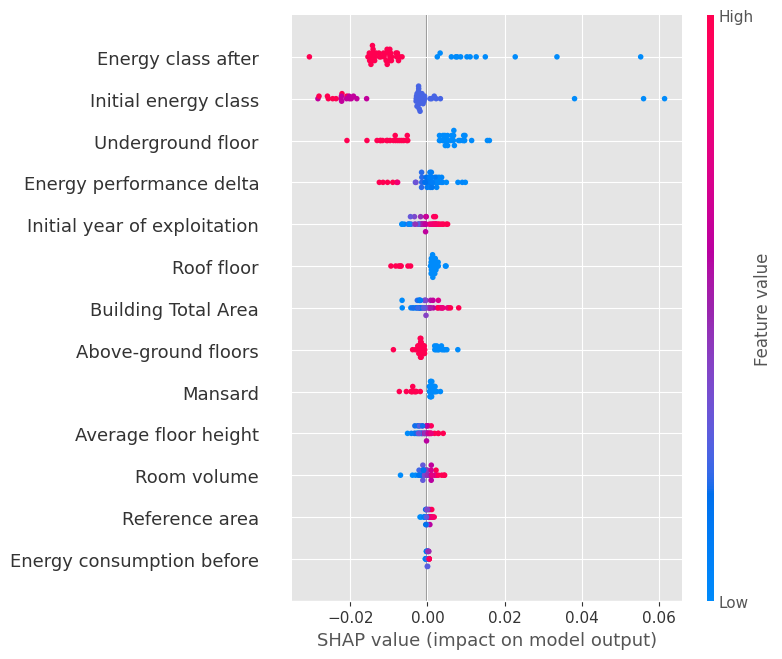}
        \caption{Heating System Installations}
        \label{fig:sum_heat_new}
    \end{subfigure}
    \caption{Summary plots of SHAP values for each class including the \textit{Energy performance delta} feature in the Latvian dataset}
    \label{fig:sum_plots_new}
\end{figure}

\newpage
\section{Waterfall plots}
\begin{figure}[h!]
    \centering
    \begin{subfigure}[b]{0.495\textwidth}
        \centering
        \includegraphics[width=\linewidth]{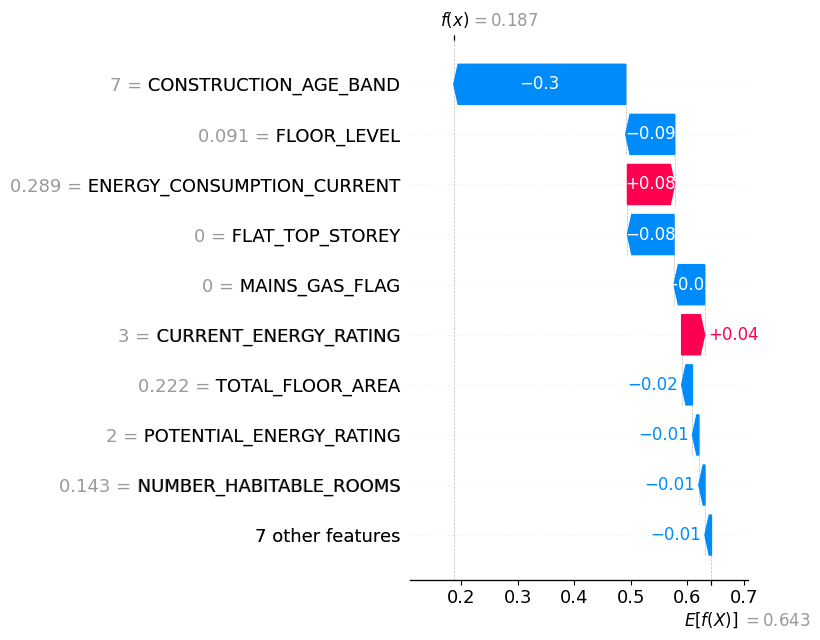}
        \caption{Building Fabric Interventions}
        \label{fig:waterfall_carr_uk}
    \end{subfigure}
    \hfill
    \begin{subfigure}[b]{0.495\textwidth}
        \centering
        \includegraphics[width=\linewidth]{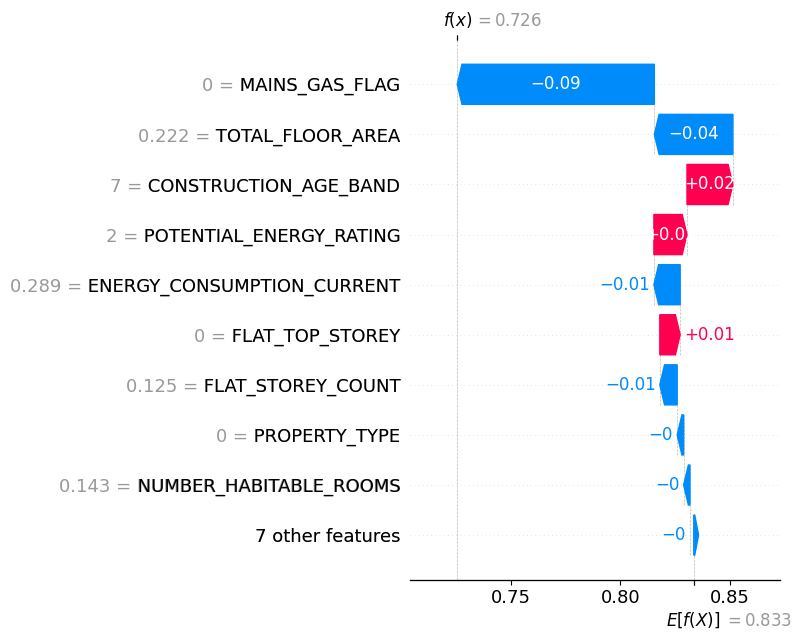}
        \caption{Heating and Lighting Controls}
        \label{fig:waterfall_rec_uk}
    \end{subfigure}
    \vspace{0.1cm}
    \begin{subfigure}[b]{0.495\textwidth}
        \centering
        \includegraphics[width=\linewidth]{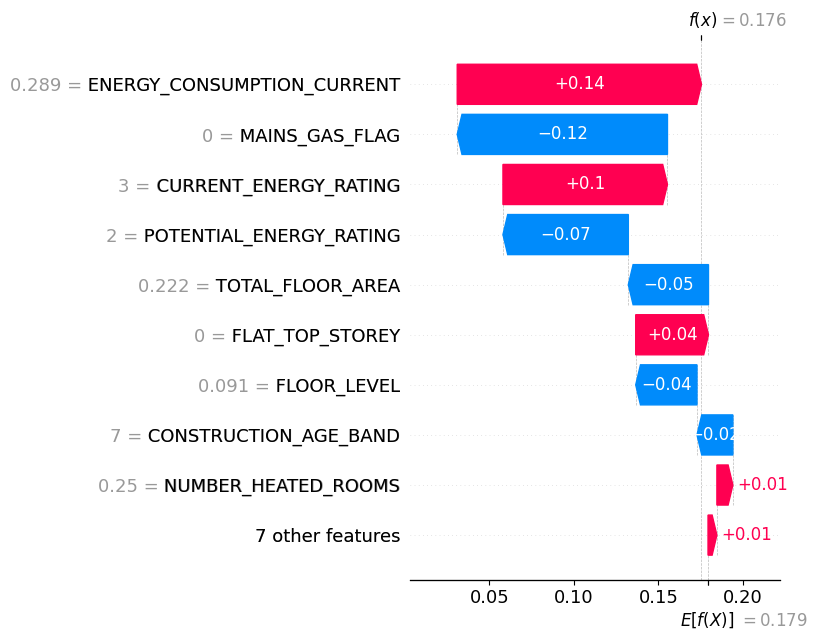}
        \caption{DHW Upgrades}
        \label{fig:waterfall_water_uk}
    \end{subfigure}
    \hfill
    \begin{subfigure}[b]{0.495\textwidth}
        \centering
        \includegraphics[width=\linewidth]{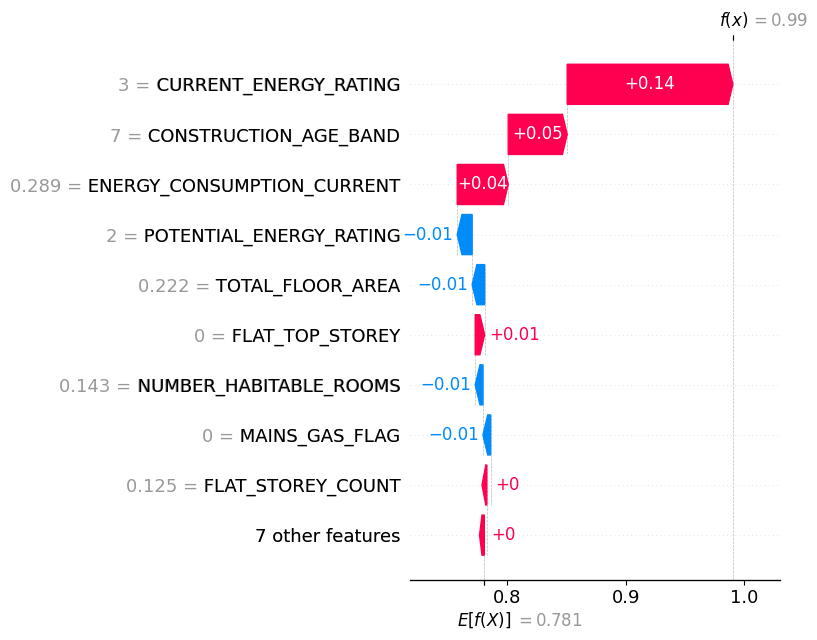}
        \caption{Heating System Installations}
        \label{fig:waterfall_heat_uk}
    \end{subfigure}
    \caption{Waterfall SHAP plots for each retrofit class of a specific indicative sample - UK dataset}
    \label{fig:waterfall_plots_uk}
\end{figure}

\begin{figure}[h!]
    \centering
    \begin{subfigure}[b]{0.495\textwidth}
        \centering
        \includegraphics[width=\linewidth]{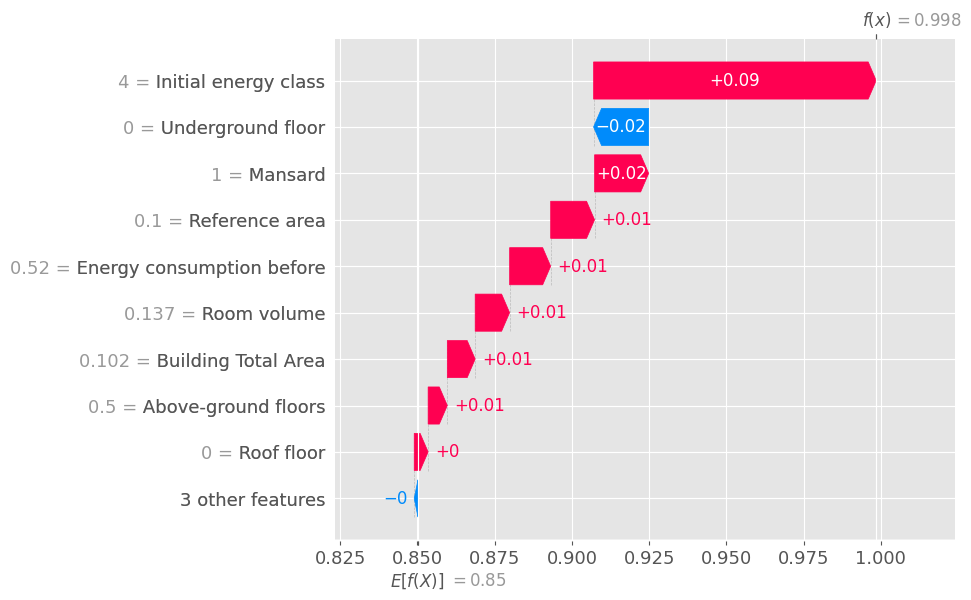}
        \caption{Building Fabric Interventions}
        \label{fig:waterfall_carr}
    \end{subfigure}
    \hfill
    \begin{subfigure}[b]{0.495\textwidth}
        \centering
        \includegraphics[width=\linewidth]{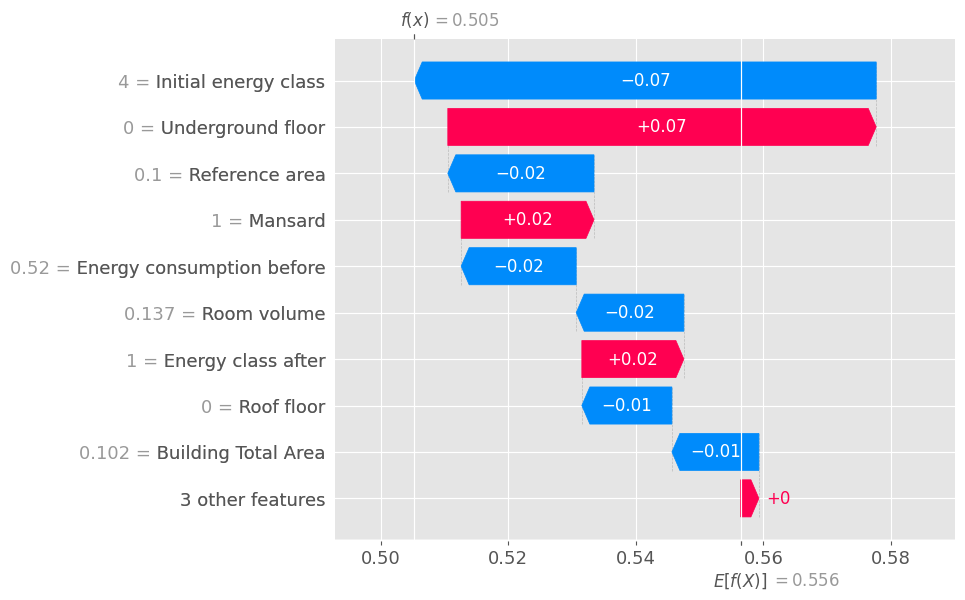}
        \caption{Heating and Lighting Controls}
        \label{fig:waterfall_rec}
    \end{subfigure}
    \vspace{0.1cm}
    \begin{subfigure}[b]{0.495\textwidth}
        \centering
        \includegraphics[width=\linewidth]{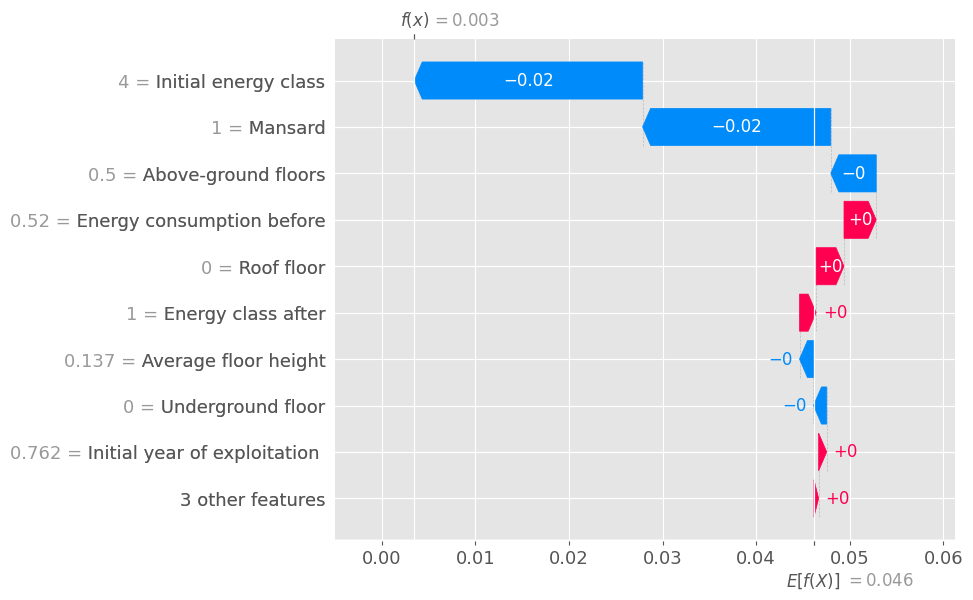}
        \caption{DHW Upgrades}
        \label{fig:waterfall_water}
    \end{subfigure}
    \hfill
    \begin{subfigure}[b]{0.495\textwidth}
        \centering
        \includegraphics[width=\linewidth]{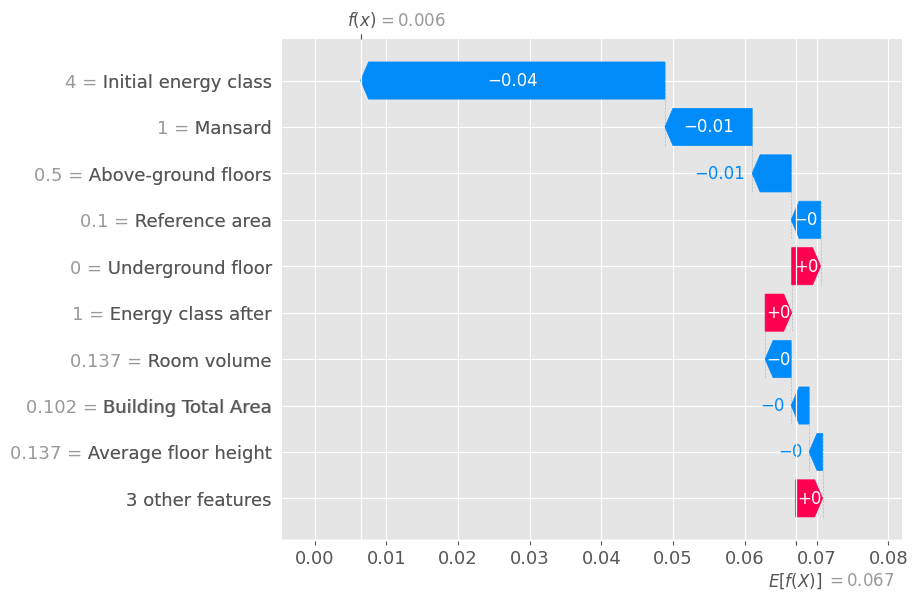}
        \caption{Heating System Installations}
        \label{fig:waterfall_heat}
    \end{subfigure}
    \caption{Waterfall SHAP plots for each retrofit class of a specific indicative sample - Latvian dataset}
    \label{fig:waterfall_plots}
\end{figure}

\newpage
\section{UK dataset: output mapping} \label{appendix_mapping}

\begin{itemize}
    \item Building Fabric Interventions: Cavity wall insulation, Draught proofing, Internal or external wall insulation, Replace single glazed windows with low-E double glazed windows, Secondary glazing to single glazed windows, Increase loft insulation to 270 mm
    \item Heating and Lighting Controls: Low energy lighting for all fixed outlets, Heating controls (programmer and TRVs), Heating controls (programmer and room thermostat), Heating controls (programmer, room thermostat and TRVs), Heating controls (room thermostat and TRVs), Heating controls (room thermostat), Heating controls (thermostatic radiator valves), Heating controls (time and temperature zone control)
    \item DHW Upgrades: Increase hot water cylinder insulation, Hot water cylinder thermostat, Insulate hot water cylinder with 80 mm jacket, Add additional 80 mm jacket to hot water cylinder
    \item Heating System Installation: Wood pellet stove with boiler and radiators, Change heating to gas condensing boiler, Change room heaters to condensing boiler, Replace boiler with new condensing boiler, Replace heating unit with condensing unit, Replacement warm air unit, Condensing boiler (separate from the range cooker), Fan assisted storage heaters, Fan assisted storage heaters and dual immersion cylinder, Fan-assisted storage heaters, Condensing oil boiler with radiators
\end{itemize}

\end{document}